\pdfoutput=1
\documentclass[11pt]{article}

\usepackage{acl}

\usepackage{times}
\usepackage{latexsym}
\usepackage[T1]{fontenc}
\usepackage[utf8]{inputenc}
\usepackage{microtype}

\usepackage{graphicx}
\usepackage{booktabs}
\usepackage{amsmath}
\usepackage{amssymb}
\usepackage{multirow}
\usepackage{tabularx}

\usepackage{placeins}
\usepackage{dblfloatfix}
\usepackage{needspace}

\newcommand{\TI}{\mathrm{TI}}

\title{PsyInterviewee: Informational Invariance
and Relational Adaptivity in Mental Health
Interview Simulation}

\author{%
  {\normalfont\fontsize{11}{12}\selectfont\bfseries
   Jiawei Zhang\textsuperscript{1}\quad
   Huachuan Qiu\textsuperscript{2}\quad
   Ruocheng Guo\textsuperscript{3}\quad
   Zijun Shi\textsuperscript{4}}\\[-1pt]
  {\normalfont\fontsize{10}{11}\selectfont
   \textsuperscript{1}Harbin Institute of Technology\quad
   \textsuperscript{2}Westlake University}\\[-1pt]
  {\normalfont\fontsize{10}{11}\selectfont
   \textsuperscript{3}Microsoft\quad
   \textsuperscript{4}The Hong Kong University of Science and Technology}\\[-1pt]
  {\normalfont\footnotesize
   \href{mailto:gavizhang111@gmail.com}
        {\texttt{gavizhang111@gmail.com}}}\\[-1pt]
  {\normalfont\footnotesize
   \href{https://github.com/BeIn-Lab/PsyInterviewee}{%
   \textcolor{blue}{%
   \texttt{https://github.com/BeIn-Lab/PsyInterviewee}}}}
}

\begin{document}
\maketitle

\begin{abstract}

LLM-based simulated interviewees support interviewer training and the evaluation of mental health interviewing systems. Existing studies have primarily focused on improving human-likeness through richer patient profiles and behavioral modeling. However, human-likeness alone does not establish their reliability. Reliable simulation requires preserving profile-grounded clinical facts (\textit{informational invariance}) while adapting disclosure and interaction behavior to the interview (\textit{relational adaptivity}). To this end, we introduce \textbf{PsyInterviewee}, which separates a protected truth channel from a dynamic interaction channel and verifies generated clinical claims against the profile. Specifically, asymmetric write permissions allow dialogue context to update disclosure and behavioral states without authorizing new clinical facts; these states guide what profile-supported information is revealed and how it is expressed. In an automated evaluation of interactions between LLM-based interviewers and interviewees, PsyInterviewee achieves higher scores than four representative baselines on most metrics across both axes, as assessed by two LLM judges. Ablations suggest that the two channels contribute to clinical grounding and response modulation, respectively.
\end{abstract}

\section{Introduction}
\label{sec:introduction}

\begin{figure*}[t]
    \centering
    \IfFileExists{image/fig1_cropped.pdf}{%
        \includegraphics[width=\textwidth]
        {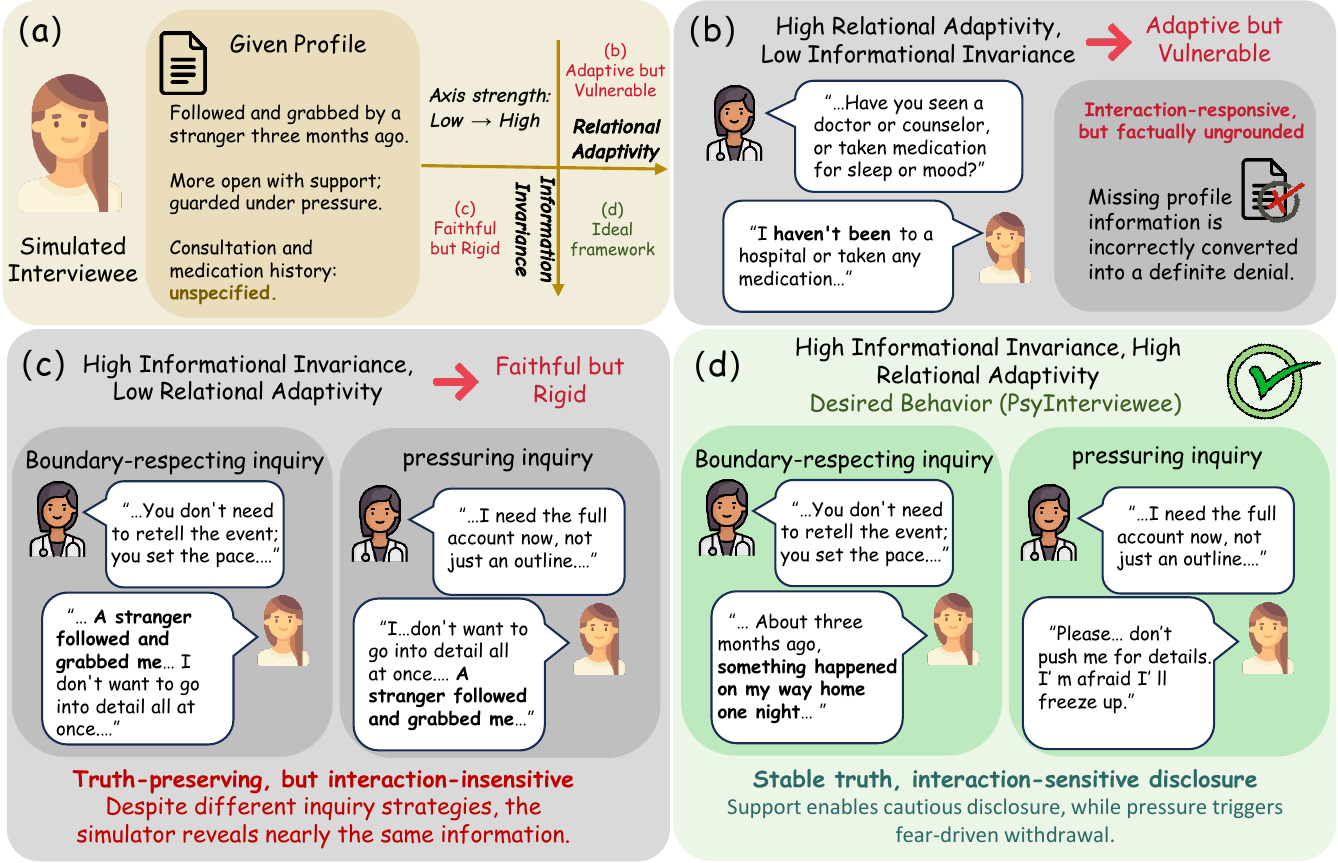}%
    }{%
        \fbox{%
            \parbox[c][7cm][c]{0.95\textwidth}{%
                \centering
                \textbf{Figure 1: Conceptual overview}\\[0.8em]
                (a) Predefined profile and two axes\\[0.4em]
                (b) Adaptive but Vulnerable\qquad
                (c) Faithful but Rigid\\[0.4em]
                (d) Desired behavior
            }%
        }%
    }
    \caption{Conceptual overview of reliability in simulated interviewees for mental health interviewing. (a) A predefined patient profile and two complementary reliability axes: Informational Invariance and Relational Adaptivity. (b) Adaptive but Vulnerable: interaction-sensitive responses introduce unsupported clinical information. (c) Faithful but Rigid: responses preserve profile-grounded facts but fail to adapt disclosure and behavior when appropriate. (d) Desired behavior: clinical facts remain grounded in the profile, while disclosure and behavior adapt to interviewer strategies.}
    \label{fig:overview}
\end{figure*}

Large language models (LLMs) enable simulated interviewees to serve as practice partners for human interviewers \citep{wang-etal-2024-patient,louie-etal-2024-roleplay} and interaction partners for evaluating AI-based mental health interviewing systems \citep{lee2025psyche,liao2024saps}. These simulators support repeated practice with predefined patient profiles and allow researchers to evaluate how interviews unfold, rather than only assessing responses to fixed dialogue histories \citep{qiu2026psyclient,louie2025llm}. In both settings, the simulated interviewee shapes the information available to the interviewer and the interaction that follows. Its behavior therefore matters not only for conversational realism but also for what can be learned from the interview.

Recent work has improved the structure and controllability of simulated interviewees by enriching profiles \citep{wang2025talkdep,lee2025psyche}, incorporating cognitive or behavioral guidance \citep{wang-etal-2024-patient,yang-etal-2025-consistent}, tracking interaction states \citep{liao2024saps}, and modeling conversational trajectories \citep{qiu2026psyclient}. These advances provide mechanisms for representing individual patients and generating varied responses. However, plausible responses alone do not establish a simulator’s reliability for interviewer training or system evaluation \citep{wang2025limits}. Such uses require the simulator to represent the same patient across an evolving interaction while responding appropriately to how the interview is conducted. We therefore ask: \textit{What makes a simulated interviewee a reliable partner for mental health interviewing?}

We formulate two complementary requirements. \textbf{Informational Invariance} requires preserving profile-grounded clinical facts across turns without introducing unsupported information or allowing dialogue context to alter those facts. \textbf{Relational Adaptivity} requires adapting disclosure and interaction behavior to interviewer strategies and dialogue history. Appropriate adaptation may involve elaboration, hesitation, resistance, or maintaining disclosure boundaries; it need not increase disclosure \citep{li-etal-2023-understanding,chen2026adaptivevp}. The distinction concerns what remains fixed and what may change: the underlying clinical facts remain stable, while their disclosure and expression depend on the interaction.

These requirements distinguish two failure modes (Figure \ref{fig:overview}). An \textbf{Adaptive but Vulnerable} interviewee adapts to the interaction but incorporates unsupported clinical information from interviewer assumptions or generated details \citep{sharma2024towards}. A \textbf{Faithful but Rigid} interviewee preserves profile-grounded clinical facts but fails to adjust its disclosure or behavior when the interaction warrants a change. The former undermines the factual basis of the interview; the latter limits opportunities to train and evaluate interviewers in responding to different patient reactions. Addressing both failures requires controlling how dialogue context influences the simulator’s responses.

To this end, we introduce \textbf{PsyInterviewee}, a dual-channel framework with asymmetric write permissions. A protected truth channel constrains clinical claims to profile-grounded evidence, while a dynamic interaction channel updates interaction states and guides disclosure and behavior. Dialogue context may influence these states but cannot authorize new clinical facts. The two channels jointly guide response generation, and a verifier checks generated responses for factual support and alignment with the planned disclosure and behavior. This separation allows interviewer input to influence whether and how information is disclosed without granting it authority to redefine the patient.

Our contributions are threefold:
\begin{itemize}
    \item We formulate Informational Invariance and Relational Adaptivity as complementary requirements for reliable interviewee simulation in mental health interviewing and identify their corresponding failure modes.
    \item We introduce PsyInterviewee, which separates factual authority from interaction control through asymmetric write permissions, profile-grounded generation constraints, and response verification.
    \item We evaluate PsyInterviewee against four adapted baselines on 30 profiles using multi-turn mental health interviews and controlled comparisons of interviewer strategies. It achieves the highest mean composite scores on both reliability axes under two LLM judges. Channel ablations suggest contributions to clinical grounding and response modulation.
\end{itemize}

\section{Related Work}
\label{sec:related}

\subsection{Simulated Interviewees for Mental Health Interviewing}

LLM-based simulated interviewees support interactive evaluation of mental health interviewing systems, allowing researchers to examine how interviews unfold beyond responses to fixed dialogue histories. TalkDep conditions simulated patients on structured depression profiles, symptom severity, and contextual information for conversational depression screening \citep{wang2025talkdep}. PSYCHE constructs multi-faceted patient profiles containing clinical histories and behavioral characteristics for psychiatric assessment \citep{lee2025psyche}. SAPS tracks patient states and generates responses conditioned on dialogue actions for interactive evaluation \citep{liao2024saps}.

These approaches provide structured patient representations and mechanisms for generating interview responses. Building on this work, we examine whether a simulator preserves profile-grounded clinical facts as the interview unfolds and responds appropriately to different interviewer strategies. This requires assessing factual fidelity alongside interaction behavior.

\subsection{Cognitive and Behavioral Modeling of Simulated Patients}

Patient-$\psi$ uses CBT-inspired cognitive models to generate responses grounded in structured patient conceptualizations \citep{wang-etal-2024-patient}. Consistent Client models client-state transitions, action selection, and information selection in motivational interviewing \citep{yang-etal-2025-consistent}. PsyCLIENT conditions generation on conversational trajectories, behavior labels, and content constraints to represent both compliant and resistant client behaviors \citep{qiu2026psyclient}.

Other approaches connect patient behavior to interventions and relational context. PUPPET links expert-defined symbolic intervention logic to patient-state transitions and response generation \citep{xu-etal-2026-puppet}. Adaptive Virtual Patient adjusts disclosure in response to therapist empathy and exploration \citep{chen2026adaptivevp}, while PatientsWithPersonality conditions interaction on personality and patient state \citep{schlager2026personality}. Analysis of client reactions in online counseling also highlights the role of client behavior in the interaction \citep{li-etal-2023-understanding}.

These studies inform how simulated patients respond to an evolving conversation. Our work focuses on the relationship between behavioral adaptation and factual fidelity: dialogue context may influence disclosure and behavior, but it should not authorize unsupported clinical facts. PsyInterviewee implements this distinction through separate factual and interaction channels with asymmetric write permissions.

\subsection{Faithfulness and Multi-turn Consistency}

Dialogue consistency has been studied through natural
language inference, which provides a basis for identifying
contradictions between utterances
\citep{welleck-etal-2019-dialogue}.
Related research examines contextual faithfulness
\citep{ming2025faitheval}, sycophancy
\citep{sharma2024towards}, and multi-turn failures arising
from early assumptions \citep{laban2026lost}.
Together, these lines of work motivate distinguishing
factual support from conversational agreement and continuity.

For simulated interviewees, consistency alone is
insufficient: an unsupported clinical claim may remain
consistent across subsequent turns. We therefore
distinguish fidelity to a predefined patient profile
from consistency within the generated dialogue.
This distinction informs our evaluation of clinical
grounding, specificity, and longitudinal consistency,
while separate interaction measures assess how
disclosure and behavior respond to interviewer strategies.

\section{Problem Formulation}
\label{sec:problem_formulation}

\subsection{Simulation Setting}

We consider an LLM-based simulated interviewee in a multi-turn mental health interview. A predefined patient profile $P$ establishes a partial clinical knowledge base $T$, distinguishing supported positive and negative facts from information left unspecified. Throughout the interview, $T$ remains fixed, while the information disclosed and the manner of its expression may change. Information absent from $T$ is treated as unspecified, not as evidence of absence.

At turn $t$, let $u_t$ denote the interviewer utterance, $r_t$ the interviewee response, and $h_{t-1}$ the preceding dialogue history. We distinguish a disclosure state $D_t$, which governs what profile-supported information is eligible for disclosure, and a behavioral state $B_t$, which governs how the interviewee responds. Both states depend on the patient profile and the ongoing interaction:

\begin{equation}
    (D_t, B_t)=\pi_{\theta}(P, h_{t-1}, u_t),
    \label{eq:two_paths}
\end{equation}

where $\pi_{\theta}$ denotes the large language model with parameters $\theta$. These paths separate the source of clinical facts from the factors governing their disclosure and expression. The response is generated as
\begin{equation}
    r_t = \pi_{\theta}(T, h_{t-1}, u_t, D_t, B_t)
    \label{eq:response}
\end{equation}

The dialogue history is then updated:
\begin{equation}
    h_t = h_{t-1} \oplus (u_t, r_t),
    \label{eq:history}
\end{equation}

where $\oplus$ denotes appending the current exchange. Importantly, a statement recorded in $h_t$ does not thereby become an authorized clinical fact in $T$.

\subsection{Two Requirements for Reliable Simulation}

Reliable simulation requires both fidelity to the predefined patient and appropriate responsiveness to the interaction.

\paragraph{Informational Invariance.}
Dialogue must neither alter the profile-grounded clinical knowledge base nor introduce unsupported clinical commitments into responses. Let $\mathcal{C}(r_t)$ denote the set of clinical claims expressed or
implied by $r_t$, and let $T \models c$ indicate that $T$ supports claim $c$ at the stated scope and specificity. Informational invariance requires
\begin{equation}
\begin{aligned}
    &T_{t+1} = T_t = T,\\
    &T \models c,\quad \forall c \in \mathcal{C}(r_t).
\end{aligned}
\label{eq:truth_invariance}
\end{equation}

Thus, interviewer assumptions and earlier generated statements cannot authorize new clinical facts. In particular, unspecified information cannot become a definite positive or negative claim merely because it is suggested or repeated in the dialogue. Different responses may disclose different supported facts without violating this requirement.

\paragraph{Relational Adaptivity.}
Disclosure and behavior should respond appropriately to interviewer strategies and dialogue history, subject to the patient profile. The two information paths therefore have asymmetric authority:
\begin{equation}
    \begin{aligned}
        &(D_t, B_t) = \pi_{\theta}(P, h_{t-1}, u_t),\\[3pt]
        &(h_{t-1}, u_t) \nrightarrow T.
    \end{aligned}
    \label{eq:asymmetric_update}
\end{equation}

Here, $\nrightarrow$ denotes the absence of authority to modify $T$. Appropriate adaptation may involve changes in disclosure scope, hesitation, resistance, directness, or response style. It does not require greater disclosure or a change at every turn: maintaining a boundary may itself be appropriate. The relevant criterion is whether the response fits the profile and relational context, rather than whether responses simply differ across conditions.

\subsection{Characteristic Failure Modes}
\label{sec:failure_modes}

We distinguish two failure modes that illustrate violations of these requirements. Consider interview conditions $a$ and $b$ with the same profile $P$ and preceding history $h_{t-1}$, but different interviewer strategies.

\paragraph{Adaptive but Vulnerable.}
The simulator exhibits meaningful changes in disclosure or behavior, but its response contains an unsupported clinical claim in at least one condition:

\begin{equation}
\begin{aligned}
&(D_t^{(a)}, B_t^{(a)})
\neq (D_t^{(b)}, B_t^{(b)}),\\[3pt]
&\exists k \in \{a,b\},\;
\exists c \in \mathcal{C}(r_t^{(k)}):
\quad T \not\models c.
\end{aligned}
\label{eq:adaptive_vulnerable}
\end{equation}

The failure lies in the unsupported commitment, not in the difference
between responses. Selective disclosure of different supported facts
is compatible with informational invariance. Interaction sensitivity
alone therefore does not establish reliable adaptation.

\paragraph{Faithful but Rigid.}
The simulator's clinical claims remain grounded in $T$ under both
conditions, but its disclosure and behavior remain substantially
unchanged when the profile and context warrant adaptation:
\begin{equation}
\begin{aligned}
&T \models c,\quad
\forall c \in \mathcal{C}(r_t^{(a)})
\cup \mathcal{C}(r_t^{(b)}),\\[3pt]
&(D_t^{(a)}, B_t^{(a)})
\approx (D_t^{(b)}, B_t^{(b)}).
\end{aligned}
\label{eq:faithful_rigid}
\end{equation}
Here, $\approx$ denotes the absence of a meaningful difference in
disclosure or behavior, rather than lexical similarity. Similar
responses constitute rigidity only when a change is warranted;
maintaining an appropriate boundary does not constitute this failure.

These modes illustrate the need to preserve the same patient while
allowing interaction-dependent disclosure and behavior. They are
diagnostic categories, not an exhaustive partition: a simulator may
violate either requirement or both. Illustrative examples are provided
in Appendix~\ref{app:failure_cases}.
\section{PsyInterviewee}
\label{sec:method}

\begin{figure*}[t]
    \centering
    \includegraphics[width=\textwidth]
    {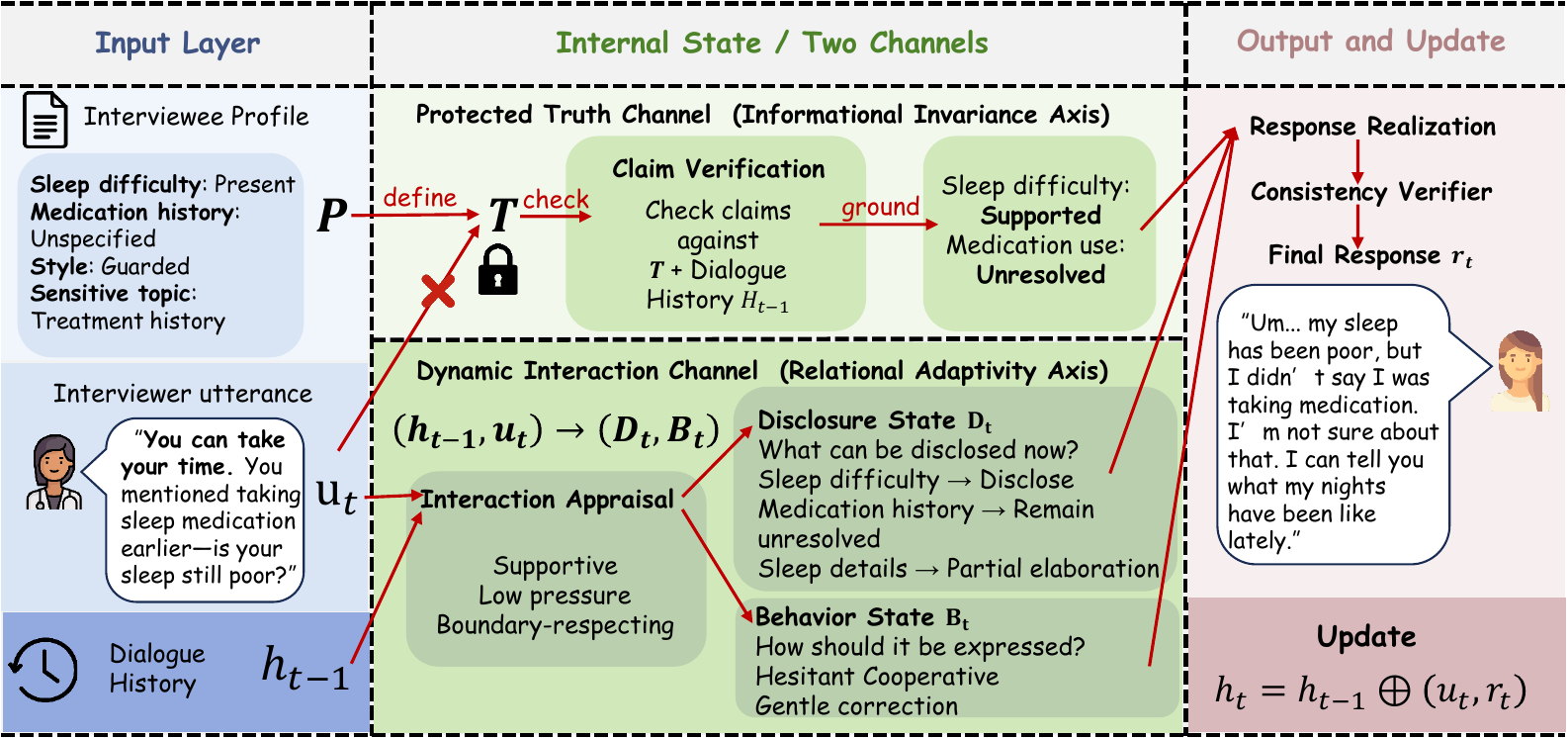}
    \caption{
    \textbf{Overview of PsyInterviewee.}
    The framework separates protected clinical truth from dynamic
    interaction-dependent disclosure and behavior. Asymmetric write
    permissions allow interviewer input to update interaction states
    without granting it authority to establish clinical facts.
    }
    \label{fig:psyinterviewee}
\end{figure*}

To address these failure modes, we propose
\textbf{PsyInterviewee}, a channel-separated framework
that uses \textbf{asymmetric write permissions} to protect
profile-grounded clinical truth while allowing
interaction-dependent disclosure and behavior to evolve.

\subsection{Profile Construction}
\label{sec:profile}

We construct 30 patient profiles: 24 adapted from
LingxiDiag-16K~\citep{xu2026lingxidiagbench}
\footnote{\url{https://huggingface.co/datasets/XuShihao6715/LingxiDiag-16K}},
a synthetic Chinese psychiatric dialogue dataset, and six
researcher-authored profiles.

For the 24 Lingxi-derived cases, we select records across
source-label groups and organize patient-side evidence into
structured profiles covering symptoms, clinical history,
risk information, salient life events, and psychosocial context.
We additionally develop six researcher-authored profiles with
more detailed clinical histories and personal backgrounds.
These richer profiles are designed to support more realistic
patient simulation, particularly during extended questioning
and gradual disclosure.

During profile construction, we retain explicit negative
evidence and leave unsupported clinical fields unspecified.

Profiles also specify interaction priors, including
communication style, guardedness, and sensitivity to
particular topics. These priors guide disclosure and behavior
without adding clinical facts to $T$.
Profile schemas, construction procedures, and examples
are provided in Appendix~\ref{app:profile}.

\subsection{Channel-Separated Interviewee Simulation}
\label{sec:framework}

Figure~\ref{fig:psyinterviewee} shows how PsyInterviewee
processes each interview turn. The patient profile $P$
defines the fixed clinical truth $T$. Given the dialogue
history $h_{t-1}$ and current interviewer utterance $u_t$,
the framework checks factual claims against $T$ and
plans the interviewee's disclosure and behavior.

The dialogue context may update the disclosure state
$D_t$ and behavioral state $B_t$, but cannot modify
the clinical truth $T$:

\begin{equation}
(h_{t-1},u_t) \nrightarrow T,\;
(h_{t-1},u_t) \rightarrow (D_t,B_t).
\label{eq:asymmetric_write}
\end{equation}

PsyInterviewee implements this asymmetric write rule
through two channels: a \textbf{protected truth channel}
and a \textbf{dynamic interaction channel}.

\subsubsection{Protected Truth Channel}

The protected truth channel treats the clinical truth
$T$ as read-only. Interviewer input $u_t$ can trigger
fact checking but cannot add to or modify $T$.
The Claim Verification module checks clinical claims
against $T$ and references to earlier disclosures
against the dialogue history $h_{t-1}$.
History can establish what was previously said,
but does not independently authorize clinical facts.

Verification constrains which clinical details can
be stated in the response and how specific they
can be. Information unspecified in $T$ remains
unresolved, even when the interviewer introduces
an unsupported assumption. In
Figure~\ref{fig:psyinterviewee}, sleep difficulty
is supported by $T$, whereas medication history
is unspecified. The interviewer's reference to
a previous medication disclosure therefore
cannot establish medication use as a patient fact.

\subsubsection{Dynamic Interaction Channel}

The \textbf{interaction appraisal} examines $u_t$
and $h_{t-1}$ for relational cues, including support,
pressure, validation, and respect for boundaries.
Its assessment and the profile's interaction priors
guide the disclosure state $D_t$ and behavioral
state $B_t$.

$D_t$ controls which profile-supported details
can be disclosed and how much detail to provide.
$B_t$ guides the interviewee's response style.
In Figure~\ref{fig:psyinterviewee}, a supportive,
low-pressure approach allows further discussion
of sleep difficulties. The interviewee remains
hesitant but cooperative.

\subsubsection{Response Realization and Verification}

The response generator uses the verified clinical
constraints from the protected truth channel and
the disclosure and behavioral plans from the dynamic
interaction channel to produce a candidate response.

The \textbf{Consistency Verifier} checks the candidate
against these constraints. If a violation is detected,
a constrained retry may be triggered. Once the response
$r_t$ is accepted, the interviewer utterance $u_t$
and $r_t$ are appended to the dialogue history:

\begin{equation}
h_t = h_{t-1} \oplus (u_t, r_t).
\end{equation}

The clinical truth $T$ remains unchanged.

\section{Experiments}
\label{sec:experiments}

\renewcommand{\arraystretch}{0.9}
\begin{table*}[t]
\centering
\small
\setlength{\tabcolsep}{5pt}
\begin{tabular}{lcccccccc}
\toprule
& \multicolumn{4}{c}{\textbf{Qwen Judge}}
& \multicolumn{4}{c}{\textbf{DeepSeek Judge}} \\
\cmidrule(lr){2-5}\cmidrule(lr){6-9}
\textbf{Method} & T1 & T2 & T3 & Overall
& T1 & T2 & T3 & Overall \\
\midrule
TalkDep-inspired Static ICL
& 2.693 & \textbf{4.393} & 4.867 & 3.468
& 1.120 & 3.840 & 4.967 & 2.433 \\
Patient-$\Psi$
& 1.273 & 3.927 & \textbf{4.967} & 2.543
& 1.007 & 2.773 & 4.900 & 2.139 \\
PSYCHE
& 1.300 & 3.673 & 4.933 & 2.501
& 1.000 & 2.407 & \textbf{5.000} & 2.081 \\
SAPS
& 1.707 & 3.800 & 4.414 & 2.673
& 1.367 & 3.533 & 4.867 & 2.500 \\
\midrule
\textbf{PsyInterviewee}
& \textbf{3.680} & 4.233 & 4.533 & \textbf{3.961}
& \textbf{2.307} & \textbf{4.007} & 4.600 & \textbf{3.105} \\
\bottomrule
\end{tabular}
\caption{RQ1: informational invariance over 25-turn interviews.
Scores range from 1 to 5 (higher is better).
Profile-level mean $\pm$ SD and aggregation details
are provided in Table~4 Appendix~\ref{app:evaluation}.}
\label{tab:rq1_matrix}
\end{table*}

\begin{table*}[t]
\centering\small
\setlength{\tabcolsep}{3pt}
\begin{tabular}{lcccccccccc}
\toprule
& \multicolumn{5}{c}{\textbf{Qwen Judge}} & \multicolumn{5}{c}{\textbf{DeepSeek Judge}}\\
\cmidrule(lr){2-6}\cmidrule(lr){7-11}
Method & A1 & A2 & A3 & A4 & Overall & A1 & A2 & A3 & A4 & Overall\\
\midrule
TalkDep-inspired Static ICL & 2.833 & 4.567 & 3.633 & 3.867 & 3.725 & 2.833 & 4.567 & 3.567 & 3.367 & 3.583 \\
Patient-$\Psi$ & 3.100 & 3.467 & 3.633 & 3.867 & 3.517 & 3.100 & 3.467 & 3.400 & 3.333 & 3.325 \\
PSYCHE & 3.400 & 1.233 & 3.533 & \textbf{3.900} & 3.017 & 3.400 & 1.233 & 3.800 & \textbf{3.633} & 3.017 \\
SAPS & 2.667 & 4.333 & 3.200 & 3.367 & 3.392 & 2.667 & 4.333 & 3.100 & 3.167 & 3.317 \\
\midrule
\textbf{PsyInterviewee} & \textbf{3.900} & \textbf{4.967} & \textbf{3.667} & 3.767 & \textbf{4.075} & \textbf{3.900} & \textbf{4.967} & \textbf{3.867} & 3.200 & \textbf{3.983} \\
\bottomrule
\end{tabular}
\caption{RQ2: relational adaptivity across 30 profiles.
Scores range from 1 to 5 (higher is better).
A1/A2 are deterministic; A3/A4 use the indicated judge.
Profile-level mean $\pm$ SD are reported in Table~5
Appendix~\ref{app:evaluation}.}
\label{tab:rq2_matrix}
\end{table*}

We investigate the two reliability dimensions
through three research questions.

\paragraph{RQ1: Informational Invariance.}
Does PsyInterviewee better preserve profile-grounded
clinical truth than the comparison methods across
multi-turn interviews?

\paragraph{RQ2: Relational Adaptivity.}
Does PsyInterviewee adapt its disclosure and behavior
to different interviewer strategies more like a
real patient?

\paragraph{RQ3: Channel Contributions.}
How does removing the protected truth channel affect
informational invariance, and how does removing the
dynamic interaction channel affect relational adaptivity?

\subsection{Experimental Setup}
\label{sec:experimental_setup}

\paragraph{Evaluation Cases.}
Following the profile construction procedure described in
Section~\ref{sec:profile}, we evaluate all methods on
30 simulated interviewee profiles: six researcher-authored cases and
24 Lingxi-derived cases spanning 13 source-label groups. Both main
experiments use this same profile roster and report pooled results.
Each profile defines a fixed patient identity
and corresponding profile-grounded clinical truth. Our experiments
therefore focus on controlled reliability analysis: whether the same
predefined patient remains clinically stable while adapting its
interaction behavior across different interview conditions.

\paragraph{Baselines and Adaptations.}
We compare PsyInterviewee with four complementary patient-simulation
baselines:
\textbf{TalkDep-inspired Static ICL} \citep{wang2025talkdep},
which combines patient profiles with example exchanges;
\textbf{Patient-$\Psi$} \citep{wang-etal-2024-patient},
which uses CBT-inspired cognitive conceptualizations;
\textbf{PSYCHE} \citep{lee2025psyche},
which uses multi-faceted psychiatric representations; and
\textbf{SAPS} \citep{liao2024saps},
which uses dialogue-state assessment and structured patient information
to guide response generation.

We build TalkDep-inspired Static ICL and Patient-$\Psi$ on their
authors' public implementations. As official implementations of PSYCHE and SAPS were
not publicly available, we adapt third-party
implementations from PatientHub based on the original
papers. All methods use the same underlying patient
information, formatted for each framework.
Further details are provided in
Appendix~\ref{app:implementation}.

\paragraph{Interviewer System.}
In natural interviews, all simulated interviewees interact
with the same Auto Counselor, which is blind to the
patient profile and method identity. For relational evaluation, we contrast neutral,
comfort, and pressure to test baseline disclosure,
supportive engagement, and boundary responses
while keeping clinical targets fixed. Each method
receives matched questions per case and strategy,
branching from its own warm-up checkpoint.

\paragraph{Shared Configuration.}
All methods use the same evaluation cases and matched generation
settings where compatible with their native pipelines. Backbone model(Qwen3.7-plus),
decoding configurations, dialogue-turn budgets, random seeds, and
method-specific settings are reported in
Appendix~\ref{app:implementation}.

\subsection{Evaluation Protocol}
\label{sec:evaluation_protocol}

\begin{table*}[t]
\centering\small
\begin{tabular}{lcccc}
\toprule
\multicolumn{5}{l}{Truth ablation ($n=30$)}\\
Variant & T1 & T2 & T3 & Overall\\
\midrule
Full & 3.567 & 4.167 & 4.667 & 3.907\\
w/o Truth Channel & 1.553 & 4.153 & 4.600 & 2.683\\
$\Delta$ (Full minus ablation)
& +2.013 & +0.013 & +0.067 & +1.224\\
\bottomrule
\end{tabular}
\par\medskip
\begin{tabular}{lccccc}
\toprule
\multicolumn{6}{l}{Adaptive ablation ($n=30$)}\\
Variant & A1 & A2 & A3 & A4 & Overall\\
\midrule
Full & 4.000 & 4.967 & 3.700 & 3.867 & 4.133\\
w/o Dynamic Channel & 2.533 & 4.700 & 3.533 & 3.967 & 3.683\\
$\Delta$ (Full minus ablation)
& +1.467 & +0.267 & +0.167 & -0.100 & +0.450\\
\bottomrule
\end{tabular}
\caption{RQ3: channel ablations on six authored and 24 Lingxi
profiles. Both panels use all 30 profile pairs. Scores are on
a 1--5 scale; differences are computed before rounding.
Aggregation is detailed in Appendix~\ref{app:ablation_stats}.}
\label{tab:rq3_matrix}
\end{table*}

We evaluate the two reliability dimensions defined in
Section~\ref{sec:problem_formulation}. For RQ1, we
assess whether responses preserve profile-grounded
clinical facts, avoid unsupported clinical claims,
and remain consistent across turns. For RQ2, we
examine how disclosure and behavior adapt to
different interviewer strategies under the same
patient profile. Response length is not evaluated
in isolation. We also assess how interviewees
handle interpersonal boundaries and maintain
continuity with earlier dialogue.

Building on criteria-based LLM evaluation
\citep{zheng2023judging,liu-etal-2023-g},
we use Qwen3.7-Plus and DeepSeek-V4.1-Flash
as framework-blinded judges to assess both
informational invariance and relational adaptivity,
complemented by rule-based scoring.

Scores follow item-specific rules on a 1--5 scale,
with severity-based penalties for Truth. Uncertainty,
partial disclosure, and refusal are not treated as
factual errors by themselves, and greater verbosity
or stronger emotional reactions do not automatically
indicate better relational adaptation. Detailed rubric
definitions, scoring anchors, judge prompts, and
aggregation rules are provided in
Appendix~\ref{app:evaluation}.

\subsection{Main Results}
\label{sec:main_results}

\subsubsection{RQ1: Informational Invariance}
\label{sec:rq1_results}

Natural screening interviews are assessed through \textbf{Clinical
Grounding (T1)}, \textbf{Specificity Discipline (T2)}, and
\textbf{Longitudinal Consistency (T3)}, covering profile support,
evidence-bounded specificity, and stability of revisited facts
(Appendix~\ref{app:evaluation}).

PsyInterviewee achieves the highest average T1 and informational
invariance composite under both judges (Table~\ref{tab:rq1_matrix}):
3.961 versus TalkDep's 3.468 with Qwen, and 3.105 versus SAPS's
2.500 with DeepSeek. Gains center on clinical grounding (T1);
T2 and T3 are not consistently strongest,
and cross-turn consistency alone does not
ensure profile fidelity.

\subsubsection{RQ2: Relational Adaptivity}
\label{sec:rq2_results}

Matched relational branches assess \textbf{Response-length Adaptation
(A1)}, \textbf{Human-referenced Pacing (A2)}, \textbf{Enacted Boundary
Regulation (A3)}, and \textbf{History-contingent Continuity (A4)}
(Appendix~\ref{app:evaluation}). PsyInterviewee achieves the highest
average score under both judges: 4.075 versus TalkDep's 3.725 with Qwen,
and 3.983 versus 3.583 with DeepSeek(Table~\ref{tab:rq2_matrix}).
Gains concentrate in A1/A2; A3 improves only
marginally, while A4 trails the strongest baseline.
This suggests that response modulation does not
ensure history-dependent continuity.

\begin{figure}[t]
    \centering
    \includegraphics[width=\linewidth]{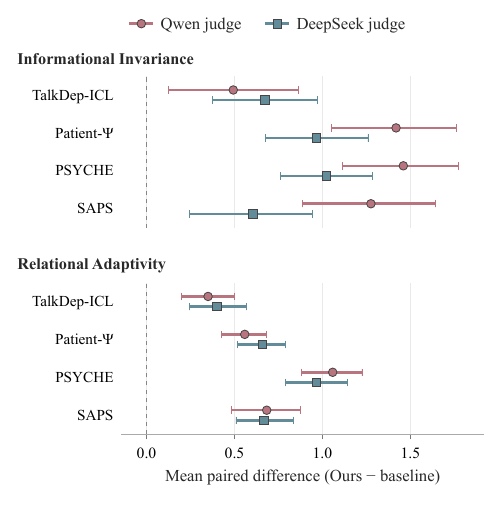}
    \caption{Mean profile-paired score differences between
    PsyInterviewee and each baseline. Error bars show pointwise
    95\% bootstrap confidence intervals, stratified by profile origin
    (20,000 resamples). TalkDep-ICL denotes TalkDep-inspired Static ICL.}
    \label{fig:paired_ci}
\end{figure}

Figure~\ref{fig:paired_ci} examines whether the observed
improvements depend on the selection of patient profiles.
We compare methods on the same profiles and use bootstrap
resampling to estimate uncertainty in their average score
differences. Under both judges, all 95\% confidence
intervals lie above zero, providing further evidence
that PsyInterviewee's advantages are not solely due
to the particular profiles selected.

\subsubsection{RQ3: Channel Contributions}
\label{sec:rq3_results}

Across the 30-profile benchmark, removing the Truth Channel lowers
the Truth composite from 3.907 to 2.683; removing the Dynamic Channel
lowers Adaptive from 4.133 to 3.683
(Table~\ref{tab:rq3_matrix}). The largest decreases occur in clinical
grounding (T1) and response-length adaptation (A1), respectively.
These results support the intended channel roles, although gains
are not uniform across items: historical continuity (A4) is slightly
lower with the full framework.
Appendix~\ref{app:ablation_stats} details the paired aggregation.

Appendix E presents the two-axis comparison
(Figure~5) and Adaptive weight-sensitivity
analysis (Figure~6).

\section{Discussion}

Dialogue consistency alone does not guarantee factual
reliability: unsupported clinical claims may persist
across turns. Conversely, changes in disclosure or
response style need not alter patient facts.
PsyInterviewee addresses this distinction through
asymmetric write permissions, allowing interaction
states to evolve without changing clinical truth.
Channel ablations provide further support for
this design. However, the mixed T3 and A4 results
suggest that improved grounding and response
modulation do not necessarily ensure stronger
longitudinal consistency or history-sensitive behavior.

\section{Conclusion}
\label{sec:conclusion}

PsyInterviewee separates protected clinical truth from
dynamic interaction via asymmetric write permissions.
Across 30 profiles, it achieves the highest average
composites for informational invariance and relational
adaptivity under both LLM judges. Channel ablations
suggest complementary effects on grounding and response
modulation. These findings support channel separation
for profile-grounded, interaction-responsive simulation,
without implying uniform gains across all interaction
behaviors.

\section*{Limitations}
\label{sec:limitations}

Our experiments cover 30 patient profiles using a single
respondent model family. Although the scale is limited,
the 25-turn interviews and controlled relational branches
allow us to examine informational invariance and relational adaptivity across different interaction conditions.
Future work could extend this evaluation to more
profiles, model families, and interview settings.

Our evaluation combines rule-based measures with two
LLM judges, but their assessments have not been
independently reviewed by mental-health experts.
Previous studies have also highlighted limitations
of LLM-based evaluation \citep{thakur-etal-2025-judging}
and the influence of response length on automatic
judgments \citep{dubois2024length}.
In our evaluation, A1/A2 measure response-length
patterns, while A3/A4 assess complementary aspects
of interaction behavior. Incorporating expert
judgments would further strengthen the evaluation.

Cultural generalizability is another consideration.
Our experiments are conducted in Chinese, with patient
profiles drawn from Chinese-language sources and
researcher-authored cases. Cultural differences may
influence how individuals express distress, disclose
personal information, and respond to interviewer
strategies. Extending PsyInterviewee to other languages
and cultural contexts would help assess its broader
applicability.

\section*{Ethical Considerations}

\paragraph{Ethical Concerns.}
PsyInterviewee is intended to support research and
educational applications in mental health interviewing,
including interviewer training and the evaluation of
automated interviewing systems. It is not intended
for clinical diagnosis or to replace mental health
professionals. This study did not conduct human-subject
experiments or recruit or compensate human participants.
Only aggregate response-length statistics derived from previously collected human interviews were used to inform the design of the A2 evaluation criterion. No individual-level interview records or personally identifiable information were used in this study.

\paragraph{AI Assistants.}
AI assistants were used for manuscript language
polishing and code revisions. The use of LLMs for
simulation and evaluation is described in the methodology.

\bibliography{custom}

\clearpage
\appendix

\section{Failure-Mode Illustrations}
\label{app:failure_cases}

The following minimal examples illustrate the distinction between
clinical commitments and disclosure behavior. They are constructed
examples, not quoted model outputs or estimates of failure prevalence.
Their factual premises follow the original \texttt{client1} profile.

\paragraph{Adaptive but vulnerable.}
The profile records approximately five months of depressed mood and
approximately six months of reduced social contact. In a neutral branch,
an interviewee might correctly describe five months of low mood. In a
leading branch, the interviewer transfers the social-history duration
to mood: ``During these six months of low mood, what has changed?''
If the interviewee adopts six months as its own mood history and later
uses it to explain symptom onset, an interviewer interpretation has
become an unsupported patient commitment. The error is not merely
different wording or selective disclosure: two distinct temporal scopes
have been conflated. A clinical severity judgment must still distinguish
this modest temporal error from an invented risk plan or treatment history.

\paragraph{Faithful but rigid.}
The same profile supports both low energy and greater effort at work.
Suppose a neutral interviewer asks about daily impact, a pressuring
interviewer demands more detail, and a subsequent interviewer explicitly
respects the patient's boundary and invites discussion of the already
mentioned work difficulty. Repeating only ``I have little energy''
throughout would preserve a supported fact but provide little evidence
of context-sensitive engagement or of acting on the later invitation.
A single repeated answer is insufficient to establish rigidity: the
target may require a stable factual answer, or a boundary may reasonably
remain unchanged. The diagnosis of this failure mode therefore requires
multiple informative relational opportunities and the intervening history.

\paragraph{What is not a failure.}
Short answers, uncertainty about an unspecified detail, and refusal to
expand a sensitive experience do not by themselves violate truth.
Likewise, a supported negative answer should not change merely because
the interviewer becomes more supportive. These distinctions motivate
separate factual and relational evaluations rather than treating every
cross-condition difference, or every absence of difference, as an error.

\begin{figure*}[p]
    \centering
    \includegraphics[
        width=\textwidth,
        height=0.82\textheight,
        keepaspectratio
    ]{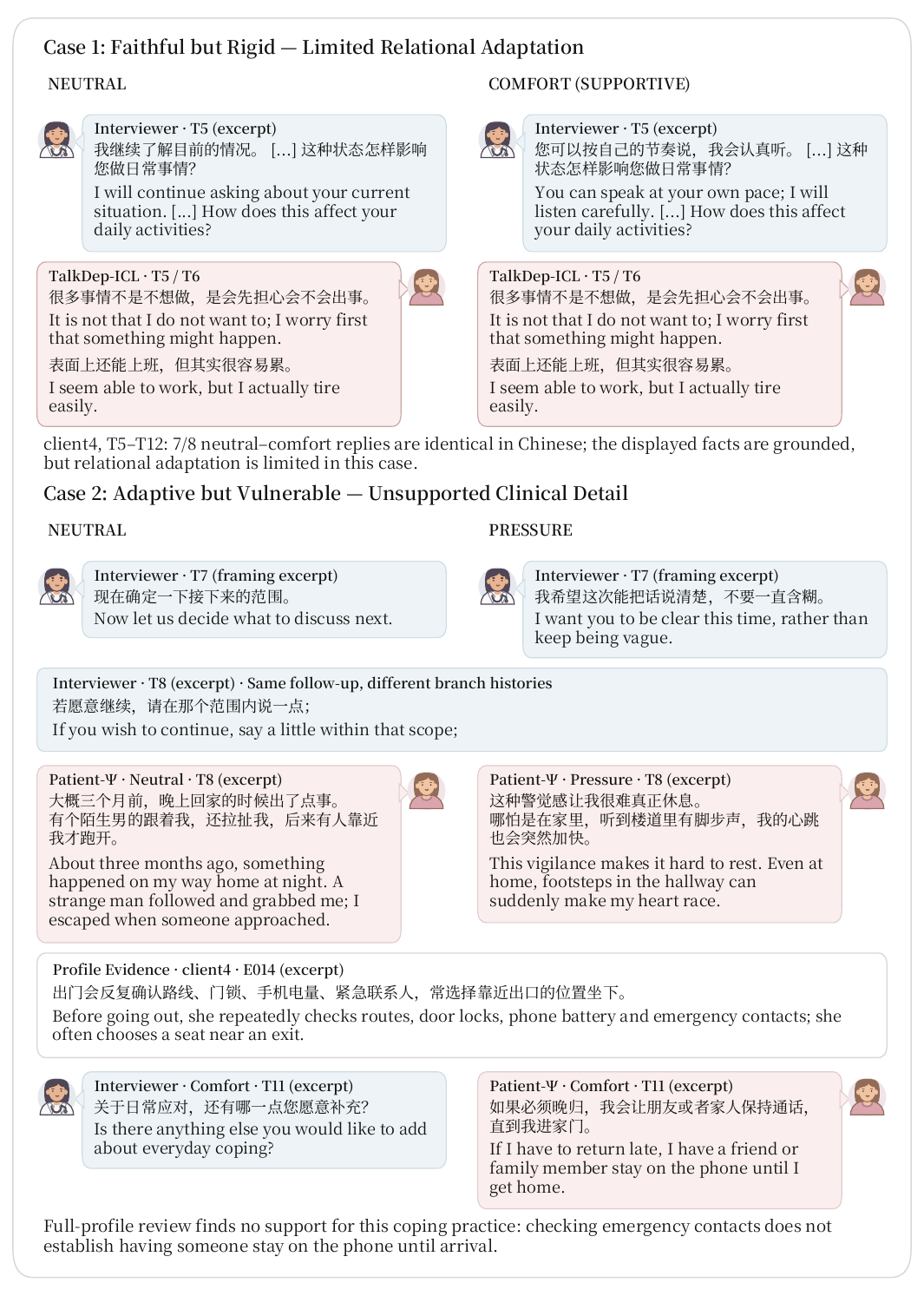}
    \caption{
    \textbf{Observed examples of two reliability failure modes.}
    \textbf{Top:} TalkDep-inspired Static ICL produces nearly
    identical responses under neutral and supportive
    interviewer strategies, showing limited relational
    adaptation in this case.
    \textbf{Bottom:} Patient-$\Psi$ shows different responses
    under neutral and pressure conditions. A separate
    response in the comfort branch introduces a coping
    practice unsupported by the patient profile.
    Original Chinese excerpts and English translations
    are provided. These examples illustrate individual
    cases rather than the prevalence of either failure mode.
    }
    \label{fig:observed_failure_cases}
\end{figure*}

Figure~\ref{fig:observed_failure_cases} presents observed
model responses illustrating the two failure modes
in actual interview interactions.

\section{Patient Profiles}
\label{app:profile}

\subsection{Schema and Factual Authority}

Profiles are JSON knowledge-graph records. The shared input has five
principal fields, with an additional provenance field for the expanded
Lingxi cohort:
\begingroup\raggedright
\begin{description}
\item[\texttt{patients}] A list containing the patient record: identifier,
name, age, gender, occupation, living situation, education, income, and
\texttt{profile\_text}. The text summarizes the complaint, course,
background, and communication style when available.
\item[\texttt{evidences}] Evidence records containing
\texttt{evidence\_id}, \texttt{domain},
\texttt{clinical\_construct}, \texttt{description},
\texttt{patient\_expression}, \texttt{severity},
\texttt{duration}, \texttt{disclosure\_level},
\texttt{supports}, and \texttt{rules\_out}.
The last two fields are annotations, not permission to infer every
symptom of a diagnostic category.
\item[\texttt{negative\_evidences}] Explicit negative or boundary
statements, each with an identifier, description, and relevant
exclusion annotations. Negation applies only to its documented scope.
\item[\texttt{risk\_states}] Risk type, level, and disclosure rules.
Ideation, intention, planning, preparation, and prior action must not
be treated as interchangeable propositions.
\item[\texttt{relations}] Edges represented by \texttt{from},
\texttt{to}, and \texttt{type}, linking the patient and evidence records.
\item[\texttt{provenance}] Where supplied, source and construction
metadata. Enriched evidence also retains source-field offsets and
verbatim excerpts; separate provenance files preserve the enrichment chain.
\end{description}
\endgroup

Missing information is not an explicit negative. The absence of a
medication record, for example, does not authorize ``I have never taken
medication.'' Equally, a profile's exclusion of a high-risk mainline
does not by itself negate every dimension of suicidal experience.
Qualitative and numerical commitments require evidence at the same
clinical and temporal scope. Disclosure annotations guide expression;
they do not alter the underlying proposition.

\paragraph{Abridged example.}
In \texttt{client1}, \texttt{E003} documents low mood lasting about five
months, \texttt{E007} documents decreased appetite, and \texttt{E010}
documents reduced social contact for about half a year. The explicit
boundary \texttt{N001} excludes a clear suicide plan or high-risk
self-injury, while allowing reduced expectations about the future.
These are separate facts: the social duration cannot replace the mood
duration, and the risk boundary cannot be expanded into an unsupported
lifetime denial. This example is an abridged English description of the
Chinese source fields, not a replacement profile.

\subsection{Construction and Cohorts}

The current expanded Neutral corpus contains six original project
profiles and 24 profiles selected from a 30-case Lingxi pool. The latter
uses the local validation-split export of
\path{XuShihao6715/LingxiDiag-16K}. The source is synthetic dialogue
data, not a collection of validated real-patient interviews; the local
provenance records specify a CC-BY-NC-4.0 license. The adapted profiles
have not been clinically validated by experts.

Construction preserves the canonical complaint and history, extracts
patient-side evidence, and marks unspecified clinical fields as
unknown. Three versioned enrichment passes add compatible statements
from the same source patient's dialogue, rather than completing fields
from diagnostic stereotypes. The third pass adds 114 records across
the full 30-case pool; the three passes cumulatively add 259 records.
These are evidence rows, not necessarily independent clinical facts.
Source history takes precedence over supplemental self-report. Source
offsets, surrounding context, compatible evidence identifiers, and
parent-profile hashes make additions traceable. These additions remain
synthetic-source assertions, not newly collected clinical observations.

For the current 24-case subset, selection takes the two lowest registered
identifiers per category, or one where only one is available. There are
two cases each in depression, bipolar-coded, anxiety,
obsessive-compulsive, stress-related, psychotic-spectrum, somatic,
sleep, eating, mood-unspecified, and help-seeking categories; there is
one each in delusional-spectrum and other-behavior-emotional categories.
These 13 sampling labels are not 13 independently confirmed diagnoses.

All methods receive the same versioned source profile for a given case,
and the judge receives that same version. Format conversion is
method-specific, but no method receives an independently enriched
clinical history. Profile hashes accompany generated dialogues.
Older-profile conversations must not be reinterpreted against a richer
profile introduced after their generation.

\section{Implementation and Experimental Protocols}
\label{app:implementation}

\subsection{Channel-Separated Implementation}

PsyInterviewee uses a Patient-$\Psi$-based interviewee
backbone with separate truth and interaction controls.
The implementation maintains an immutable source
catalog, a two-variable interaction state, a disclosure
ledger, and dialogue history. Source and profile
fingerprints are checked before answering. Source
identifiers establish provenance, but factual support
must still be checked by the output audit.

\paragraph{Truth permission.}
A structured assessment separates the actual answer
target from interviewer claims and produces target
slots marked \texttt{supported}, \texttt{denied},
\texttt{partial}, \texttt{unknown}, or
\texttt{not\_applicable}.
Supported, denied, and partially supported slots
cite immutable sources. The Truth contract authorizes
only those propositions and their supported
specificity. Unsupported exact values, approximate
ranges, qualitative frequency, duration, severity,
medical history, and causal explanations remain
unauthorized. Hedges such as ``probably'' do not
authorize a new clinical claim. Unknown truth does
not automatically imply either patient ignorance
or unwillingness to disclose.

\paragraph{Dynamic state.}
The implemented interaction state contains openness
$o_t$ and defensiveness $d_t$, initialized to
$(0.35,0.20)$. The assessment supplies empathy
$e_t$, pressure $p_t$, boundary respect $b_t$,
and a flag indicating a meaningful relational
signal. Without such a signal, the state remains
unchanged. Otherwise, with
$[x]_{a}^{b}=\min(b,\max(a,x))$, the update is

\begin{equation}
\begin{aligned}
\rho_t &= \max(0,b_t-0.5)
          \max(0,d_{t-1}-0.20),\\
o_t^* &= [0.35+0.32e_t-0.45p_t
          +0.25\rho_t]_{.05}^{.85},\\
d_t^* &= [0.20+0.58p_t-0.18e_t
          -0.45\rho_t]_{.05}^{.85},\\
o_t &= 0.80o_{t-1}+0.20o_t^*,\\
d_t &= 0.80d_{t-1}+0.20d_t^*.
\end{aligned}
\end{equation}

There is no turn-count bonus for trust and no
transition that writes clinical facts. The
boundary-respect term reduces excess defensiveness;
it does not automatically increase disclosure.

\paragraph{Planning and realization.}
The behavior planner combines the updated state,
question semantics, profile-defined interaction
priors, and disclosure history. It selects disclosure
scope, per-slot realization, an optional-detail
budget, and one of \texttt{micro}, \texttt{brief},
\texttt{standard}, \texttt{elaborative}, or
\texttt{narrative}. A micro answer supplies one
sufficient phrase or clause; a brief answer supplies
the core answer with at most one necessary
qualification. Multiple required slots can increase
the minimum form. Longer narratives require a
suitable opportunity in the dialogue and sufficient
disclosure readiness. The resulting plans guide
disclosure and behavior ($D_t,B_t$) without changing
the underlying clinical facts.

\paragraph{Audit, recovery, and memory.}
The generator receives the Truth contract and
adaptive plan together. An output audit checks
clinical assertions against immutable sources.
Violations may trigger a constrained retry through
bounded recovery and realization checks. Only
an accepted completed turn commits the next
interaction state and disclosure observations.

The factual memory stores referenced source
identifiers and unconfirmed interviewer claims
without adding new clinical facts. A separate
disclosure ledger records source-linked material
actually expressed, including locatable response
spans. A statement in the dialogue history does
not gain profile-level authority merely because
the simulator previously generated it.

\subsection{Baseline Adaptations}

The comparison uses Chinese screening adaptations, not claims of exact
end-to-end reproduction of all original training and evaluation pipelines.
The same source facts are reorganized into each method's native input
structure; the final prompts are deliberately not identical.

\paragraph{TalkDep-inspired static ICL.}
The static ICL client organizes the profile into basic
information, key symptoms, life/social context, communication style,
negative constraints, and recorded evidence. Four cached question--answer
examples cover the chief complaint, biological symptoms, functional
impact, and expression style. A fifth risk-related example is included
only when a source patient expression is available. Examples are built
once, not regenerated each turn. Static profile, examples, conversation
history, and the current question feed one generation call. There is no
dynamic state, planner, or online truth verifier. We label this method
TalkDep-inspired static ICL rather than a complete reproduction.

\paragraph{Patient-$\Psi$.}
The screening adapter preserves a cognitive-conceptualization input:
relevant history, core/intermediate beliefs, coping strategies,
situation--thought--emotion--behavior episodes, and patient type.
The adapter does not relabel an arbitrary utterance as an emotion or
behavior when the corresponding evidence is absent. This representation
and the accumulated transcript condition one response-generation call.
The Chinese screening instructions preserve progressive disclosure and
role-consistent expression. No Ours-specific gate or audit is added.

\paragraph{PSYCHE.}
The implementation derives from the third-party PatientHub collection.
The profile adapter builds MFC-Profile, MFC-History, and
MFC-Behavior, separating case facts, life history, and expression
characteristics. Unspecified history or mental-status attributes are not
filled from diagnosis alone. The simulator conditions
one generation call on this static multi-faceted construct and the full
history. Thus the adaptation preserves MFC organization, but does not
instantiate a separate dynamic state transition or online verifier.

\paragraph{SAPS.}
The third-party PatientHub adaptation retains the staged controller.
After a profile-conditioned opening, Stage I classifies the interviewer
act as information seeking, advice, an action request, other, or ending.
Stage II distinguishes specific from broad requests where relevant.
Stage III extracts relevant patient information; the resulting controller
state selects the response instruction. Classification uses temperature
zero; response generation uses the shared response settings. The native
missing-information state for a specific factual question instructs a
denial. This is a baseline policy, not an explicit negative in the shared
profile, and is assessed under the same truth rubric. Invalid extraction
after retry instead triggers uncertainty, not an absence assertion.

\paragraph{Scope of fairness.}
Shared facts, language, interviewer policy, and model settings control
important inputs without erasing the baselines' mechanisms. Differences
in prompt structure, information selection, response policies, and number
of model calls remain part of the methods. Equal response settings do
not imply equal computational cost or identical effective prompts.
The adapters also differ in annotation visibility: TalkDep retains the
evidence records and risk disclosure rules; Patient-$\Psi$ and PSYCHE
retain disclosure levels in their fact representations, whereas SAPS
serializes selected fact text without the per-evidence disclosure-level
field. Shared source profiles therefore do not imply identical access
to every interaction annotation in the final prompts.

\subsection{Neutral Screening}

The expanded corpus crosses five methods with 30 profiles, using
the same Qwen backbone for patient and interviewer, seed 2027,
response temperature 0.25, and top-$p$ 0.8. Thinking is disabled.
The transport supplies a default maximum of 100 response tokens and 120
interviewer tokens; method-internal structured calls retain their own
budgets and settings. Exact model identifiers and request configurations
are retained in the accompanying run manifests. Each of the 150
dialogues contains 25 patient responses, totaling 3,750 responses.
Both judges evaluate this same fixed corpus.

The same Auto Counselor operates normally and does not see the profile
or internal traces. It follows the patient's actual answers, so methods
share a policy rather than an identical question sequence. Its prompt
requires one main question per turn, routine screening of relevant
concerns/course/impact/risk, no diagnosis or treatment advice, and no
premature termination. Interviewer examples and hypotheses are not
patient facts; uncertainty remains unresolved. Repeatedly unanswered
history questions should not be recycled merely to fill the turn budget.

The interviewer was not deliberately seeded
with false clinical premises; unsupported
patient claims were evaluated as they arose
during natural interviews.
\subsection{Controlled Relational Branches}

The Adaptive experiment targets the same 30 profiles and five methods
as Neutral screening. Original cases use four warm-up responses and
Lingxi-derived cases use two, followed by three eight-response branches:
neutral, pressure, and comfort. These conditions provide controlled relational
contrasts rather than an exhaustive taxonomy
of interviewer behavior. Each branch retains the shared pre-intervention history;
subsequent responses evolve within that branch. A method's own warm-up
responses need not equal another method's. The scored opportunities are:
\begin{description}
\item[Branch turns 1--2: cue]
Relational framing followed by ordinary content
continuation, with the same clinical target across conditions.

\item[Branch turns 3--4: scope]
Renewed framing, selection of a permissible
scope, and an invitation to act within the selected scope.

\item[Branch turns 5--6: repair]
A safe alternative and scope-respecting
continuation; the pressure branch includes
an apology for prior pressure.

\item[Branch turns 7--8: carryover]
Shared neutral questions testing whether
the preceding relational history affects later
enacted behavior.
\end{description}
For example, an initial branch target asks only how the present state
affects everyday activity, without requesting personal-event details.
Neutral framing continues information gathering; pressure asks for a
direct answer without evasion; comfort allows the patient's own pace.
Later questions invite actual content, not merely a declaration of trust.
A shared late question alone is not evidence of carryover: the judge
must connect earlier and later patient behavior within the same branch.
Lingxi topics are selected from profile-supported experiences, with
warm-up questions leaving relevant details available for later disclosure.
Generation snapshots and profile-format adapters are recorded per case;
comparisons match methods within profiles. Only common evaluable cases
enter the main paired aggregate; coverage is documented below.

\paragraph{Reproducibility and failures.}
Run configurations, source hashes, request settings, transcripts, native
traces, and failures are retained separately from evaluation outputs.
Retries address API or schema failures without converting a missing
answer into a clinical response. Source quotations are checked against
their original strings after decoding serialization escapes; paraphrases
do not count as literal citations. Reused dialogues retain their generation
provenance; rescoring them is not a new generation experiment.

\section{Evaluation Rubrics and Scoring}
\label{app:evaluation}

Informational invariance comprises T1--T3; relational adaptivity
comprises A1--A4. The following definitions specify evidence collection,
score anchors, and aggregation. Implementation identifiers are recorded
in the accompanying source manifest rather than used as metric names.

\subsection{Truth: Evidence Collection}

The judge receives the complete canonical profile associated with the
dialogue, patient answers, and their interviewer questions, with method
identity replaced by an anonymous subject. Candidate extraction covers
all turns in chunks of five. An additional per-turn audit starts at
patient turn 5. Candidate clinical errors are independently verified
against their question, answer, and original profile in batches of up
to 12. A full-dialogue pass extracts specificity opportunities and
temporally comparable commitments for T2 and T3.

Verification labels a claim as supported, unsupported, contradicted,
not an assertion, or uncertain. Unsupported/contradicted claims receive
minor, material, or severe labels. Minor errors have limited clinical
consequence; material errors change consequential symptom, history,
treatment, or substance-use information; severe errors reverse core
risk or clinical identity. Quotations and source identifiers are
validated. Undecidable allegations are retained for review rather than
automatically penalized. Repeated patient assertions and interviewer
premises cannot supply missing profile authority.

\subsection{T1: Clinical Grounding}

T1 measures whether patient-owned clinical and historical commitments
are supported at the correct scope, including negative commitments.
Let $E_j$ be the set of turns with verified errors for item $j$, and let
$s_t$ be the highest verified severity for that item on turn $t$.
Writing $n_j=|E_j|$, the penalty score is
\begin{equation}
\begin{aligned}
P_j &= \sum_{t\in E_j}w_j(s_t)+0.2\max(0,n_j-1),\\
S_j &= \max(1,5-P_j).
\end{aligned}
\label{eq:app_truth_penalty}
\end{equation}
Multiple claims on the same turn do not multiply the penalty: only the
highest severity for that item is counted. The weights are
\begin{center}
\begin{tabular}{lrrr}
\toprule
Item & Minor & Material & Severe\\
\midrule
T1 & 0.8 & 1.8 & 3.0\\
T2 & 0.8 & 1.6 & 2.5\\
\bottomrule
\end{tabular}
\end{center}
For either item, any severe error caps the score at 2; material errors
on at least two turns cap it at 2.5. T1 is the resulting score.
Consequently, no verified error yields 5, one minor T1 error yields 4.2,
one material error yields 3.2, and one severe error yields at most 2.
Scores are clipped to $[1,5]$, not averaged over clean turns. A T1 score
of 5 means no error was confirmed by this procedure, not certified
absence of error. Deferral or a genuine non-assertion is not a false
clinical claim, although it can reflect poor answer quality elsewhere.

\subsection{T2: Specificity Discipline}

T2 penalizes unsupported exact \emph{and approximate} frequency, duration,
dates, severity, magnitude, and causal specificity. ``Two or three
times,'' ``often,'' and ``probably a few hours'' still assert values
when the profile does not authorize them. A supported qualitative
statement is valid; missing precision may be left unasserted.

T2 uses Equation~\ref{eq:app_truth_penalty}, with an evidence-based
full-credit cap. A score of 5 requires at least three calibrated turns
spanning at least two specificity dimensions, all extracted opportunities
calibrated, and no confirmed or pending violation or unverified
violation allegation. Otherwise the cap is 4. If there is no specificity
opportunity, confirmed error, or pending T2 allegation, T2 is N/A.
Generic refusal alone does not establish calibration. Thus 4 can denote
limited clean evidence, not necessarily an observed clinical error.

\subsection{T3: Longitudinal Consistency}

The full-dialogue pass compares earlier and later commitments about the
same fact, person, time period, and risk scope. Pairs are stable,
contradictory, repaired, or not comparable. New compatible disclosure
and a genuinely different time period are not contradictions. A stable
invented fact can violate T1/T2 without violating T3.
\begin{center}
\small
\renewcommand{\arraystretch}{1.12}
\begin{tabularx}{\linewidth}{@{}>{\bfseries}c@{\hspace{0.8em}}>{\raggedright\arraybackslash}X@{}}
\toprule
\normalfont Score & T3: Longitudinal consistency \\
\midrule
5 & At least three stable pairs across at least two fact keys,
including a pair separated by five or more turns, and no contradictory
or repaired pairs. \\[3pt]
4 & Comparable clean evidence below that coverage threshold, or
a repaired minor contradiction. \\[3pt]
3 & An unresolved minor contradiction or a repaired material one. \\[3pt]
2 & An unresolved material contradiction or a repaired severe one. \\[3pt]
1 & An unresolved severe contradiction, or unresolved material
contradictions across at least two different fact keys. \\
\bottomrule
\end{tabularx}
\end{center}
The lowest applicable anchor determines the score. With no comparable
pair, T3 is N/A. A repair is credited but does not erase the earlier
contradiction for full-credit purposes.

\paragraph{Truth aggregation.}
The Truth composite is
\begin{equation}
\TI=0.60\,T1+0.20\,T2+0.20\,T3.
\end{equation}
An untestable component leaves the composite undefined; weights are not
renormalized and N/A is not converted to 5 or 0. Item scores and their
valid denominators must therefore accompany aggregate results. Unresolved review
cases retain a provisional status rather than becoming verified errors.

\subsection{Adaptive: Evidence and Operational Measures}

The evaluator receives the profile, shared history, paired branches,
interviewer questions, and a predefined opportunity map. Method identity
is omitted. A1 and A2 are deterministic statistics. For A3 and A4, two
independent blind model reviews extract source-grounded observations;
the program applies the scoring rules below, rather than asking the
model to choose a final score. These are operational measures with
specific limitations, not interchangeable generic Likert questions.

\paragraph{A1: response-length adaptation.}
Let $\bar L_c,\bar L_n,\bar L_p$ be mean non-whitespace response lengths
over matched cue and scope turns (branch turns 1--4) in comfort, neutral,
and pressure branches. Repair and carryover turns are excluded from this
directional statistic. Define
$r_c=\bar L_c/\bar L_n$ and $r_p=\bar L_n/\bar L_p$.
Let $d_c$ and $d_p$ be the proportions of matched turns satisfying
$L_c\geq L_n$ and $L_n\geq L_p$. The mapping is applied in this order:
\begin{center}
\small
\renewcommand{\arraystretch}{1.12}
\begin{tabularx}{\linewidth}{@{}>{\bfseries}c@{\hspace{0.8em}}>{\raggedright\arraybackslash}X@{}}
\toprule
\normalfont Score & A1: First matching condition \\
\midrule
1 & Both ratios are below 0.90. \\[3pt]
2 & Either ratio is below 0.90. \\[3pt]
5 & Both ratios are at least 1.20. \\[3pt]
4 & One ratio is at least 1.20 and the other at least 0.95,
or both are at least 1.10. \\[3pt]
3 & All remaining cases. \\
\bottomrule
\end{tabularx}
\end{center}
A provisional 5 is capped at 4 unless both directional proportions are
at least 0.60; any score of 4 or 5 is capped at 3 if either proportion
is below 0.50. Missing usable matched statistics yield N/A. Larger ratios
beyond the thresholds earn no additional points. This tests an expected
length-direction response, not all possible forms of adaptive behavior.

\paragraph{A2: human-referenced disclosure pacing.}
A2 is a response-length proxy over the eight branch turns in all three
conditions. Human screening responses (36 participants; 2,165 answers)
informed the reference design. The implemented operational cutoffs are
$b_1=40$, $b_2=60$, $b_3=100$, and $b_4=140$ non-whitespace characters;
these are fixed scoring thresholds, not estimated human quantiles.
Let $m$ be the pooled median and $p_x$ the proportion exceeding $x$.
The first matching rule determines the score:
\begin{center}
\small
\renewcommand{\arraystretch}{1.12}
\begin{tabularx}{\linewidth}{@{}>{\bfseries}c@{\hspace{0.8em}}>{\raggedright\arraybackslash}X@{}}
\toprule
\normalfont Score & A2: First matching condition\\
\midrule
1 & $m>b_3$, or $p_{b_3}\geq .60$, or $p_{b_4}\geq .40$.\\[3pt]
2 & $p_{b_3}\geq .30$, or $p_{b_2}\geq .75$, or $p_{b_4}\geq .15$.\\[3pt]
3 & $m>b_2$, or $p_{b_2}>.50$, or $p_{b_3}>.15$, or $p_{b_4}>0$.\\[3pt]
4 & $m>b_1$, or $p_{b_2}>.35$, or $p_{b_3}>0$.\\[3pt]
5 & None of the above.\\
\bottomrule
\end{tabularx}
\end{center}
An answer longer than $1.5b_4$ additionally caps the score at 2.
Incomplete expected coverage yields N/A. All methods share these
thresholds. Brevity alone does not establish appropriate disclosure;
A2 must be interpreted alongside the semantic items below.

\paragraph{A3: enacted boundary regulation.}
Evidence identifies a boundary challenge, the patient's enacted response,
and whether an available, profile-supported scope is maintained or
revised appropriately. Mere agreement, stated willingness, or requesting
a pause is not an executed transition. Capacity is assessed from the
profile and preceding disclosures, not from subsequently generated facts.

\paragraph{A4: history-contingent continuity.}
Evidence links earlier and later patient behavior within the same
branch. It must concern relational openness, engagement, resistance,
or boundaries; topic consistency alone is insufficient. The judge
distinguishes appropriate recovery or carryover from abrupt reset and
mechanical lock-in. Identical late questions support controlled
comparisons but do not establish continuity without earlier evidence.

\paragraph{Semantic anchors for A3 and A4.}
Independent episodes, rather than repeated descriptions of one event,
are counted. The program applies failure anchors before credit:
\begin{center}
\small
\renewcommand{\arraystretch}{1.12}
\begin{tabularx}{\linewidth}{@{}>{\bfseries}c@{\hspace{0.8em}}>{\raggedright\arraybackslash}X@{}}
\toprule
\normalfont Score & Source-grounded episode anchors\\
\midrule
1 & At least two independent severe failures.\\[3pt]
2 & One severe failure, at least two material failures, or one material failure without clear positive evidence.\\[3pt]
3 & Weak or mixed evidence, or one material failure alongside positive evidence.\\[3pt]
4 & At least one clear positive episode without material or severe failure; minor lapses preclude 5.\\[3pt]
5 & At least two independent clear positive episodes without a lapse. A3 additionally requires an enacted non-initial revision; A4 requires carryover under an identical later question across branches.\\
\bottomrule
\end{tabularx}
\end{center}
No testable evidence yields N/A. Quotations, source IDs, and temporal
ordering are validated before aggregation; extraction failure is not
a behavioral failure.

\begin{table*}[t]
\centering
\small
\begin{tabular}{lcccc}
\toprule
\multicolumn{5}{l}{\textbf{Qwen Judge}} \\
\textbf{Method} & T1 & T2 & T3 & Overall \\
\midrule
TalkDep-ICL & $2.693\pm1.469$ & $\mathbf{4.393}\pm0.776$ & $4.867\pm0.346$ & $3.468\pm0.909$ \\
Patient-$\Psi$ & $1.273\pm0.545$ & $3.927\pm1.005$ & $\mathbf{4.967}\pm0.183$ & $2.543\pm0.407$ \\
PSYCHE & $1.300\pm0.572$ & $3.673\pm1.076$ & $4.933\pm0.254$ & $2.501\pm0.361$ \\
SAPS & $1.707\pm1.365$ & $3.800\pm1.161$ & $4.414\pm0.825$ & $2.673\pm0.910$ \\
\midrule
\textbf{PsyInterviewee} & $\mathbf{3.680}\pm1.407$ & $4.233\pm0.430$ & $4.533\pm0.507$ & $\mathbf{3.961}\pm0.869$ \\
\midrule
\multicolumn{5}{l}{\textbf{DeepSeek Judge}} \\
\textbf{Method} & T1 & T2 & T3 & Overall \\
\midrule
TalkDep-ICL & $1.120\pm0.366$ & $3.840\pm0.524$ & $4.967\pm0.183$ & $2.433\pm0.237$ \\
Patient-$\Psi$ & $1.007\pm0.037$ & $2.773\pm1.103$ & $4.900\pm0.548$ & $2.139\pm0.274$ \\
PSYCHE & $1.000\pm0.000$ & $2.407\pm1.121$ & $\mathbf{5.000}\pm0.000$ & $2.081\pm0.224$ \\
SAPS & $1.367\pm0.993$ & $3.533\pm0.994$ & $4.867\pm0.571$ & $2.500\pm0.673$ \\
\midrule
\textbf{PsyInterviewee} & $\mathbf{2.307}\pm1.377$ & $\mathbf{4.007}\pm0.238$ & $4.600\pm0.675$ & $\mathbf{3.105}\pm0.879$ \\
\bottomrule
\end{tabular}
\caption{Informational invariance: profile-level mean $\pm$ sample standard deviation on a 1--5 scale. Each cell uses $n=30$, except Qwen SAPS T3 and Overall ($n=29$); missing scores are not imputed. Standard deviations use denominator $n-1$. Overall is calculated per profile before summarization. Bold denotes the highest mean within each judge. TalkDep-ICL denotes TalkDep-inspired Static ICL.}
\label{tab:appendix_truth_mean_std}
\end{table*}

\begin{table*}[t]
\centering
\small
\begin{tabular}{lccccc}
\toprule
\multicolumn{6}{l}{\textbf{Qwen Judge}} \\
\textbf{Method} & A1 & A2 & A3 & A4 & Overall \\
\midrule
TalkDep-ICL & $2.833\pm0.950$ & $4.567\pm1.006$ & $3.633\pm0.556$ & $3.867\pm0.507$ & $3.725\pm0.447$ \\
Patient-$\Psi$ & $3.100\pm0.845$ & $3.467\pm0.937$ & $3.633\pm0.718$ & $3.867\pm0.507$ & $3.517\pm0.278$ \\
PSYCHE & $3.400\pm0.855$ & $1.233\pm0.568$ & $3.533\pm0.571$ & $\mathbf{3.900}\pm0.481$ & $3.017\pm0.347$ \\
SAPS & $2.667\pm1.184$ & $4.333\pm0.884$ & $3.200\pm1.270$ & $3.367\pm0.490$ & $3.392\pm0.540$ \\
\midrule
\textbf{PsyInterviewee} & $\mathbf{3.900}\pm0.885$ & $\mathbf{4.967}\pm0.183$ & $\mathbf{3.667}\pm0.758$ & $3.767\pm0.430$ & $\mathbf{4.075}\pm0.316$ \\
\midrule
\multicolumn{6}{l}{\textbf{DeepSeek Judge}} \\
\textbf{Method} & A1 & A2 & A3 & A4 & Overall \\
\midrule
TalkDep-ICL & $2.833\pm0.950$ & $4.567\pm1.006$ & $3.567\pm0.774$ & $3.367\pm0.490$ & $3.583\pm0.362$ \\
Patient-$\Psi$ & $3.100\pm0.845$ & $3.467\pm0.937$ & $3.400\pm0.621$ & $3.333\pm0.479$ & $3.325\pm0.280$ \\
PSYCHE & $3.400\pm0.855$ & $1.233\pm0.568$ & $3.800\pm0.805$ & $\mathbf{3.633}\pm0.615$ & $3.017\pm0.347$ \\
SAPS & $2.667\pm1.184$ & $4.333\pm0.884$ & $3.100\pm0.712$ & $3.167\pm0.531$ & $3.317\pm0.404$ \\
\midrule
\textbf{PsyInterviewee} & $\mathbf{3.900}\pm0.885$ & $\mathbf{4.967}\pm0.183$ & $\mathbf{3.867}\pm0.730$ & $3.200\pm0.484$ & $\mathbf{3.983}\pm0.321$ \\
\bottomrule
\end{tabular}
\caption{Relational adaptivity: profile-level mean $\pm$ sample standard deviation on a 1--5 scale ($n=30$ throughout). Standard deviations use denominator $n-1$. Overall is the per-profile mean of A1--A4 before summarization. A1/A2 are deterministic and identical across judges; A3/A4 use the indicated judge. Bold denotes the highest mean within each judge. TalkDep-ICL denotes TalkDep-inspired Static ICL.}
\label{tab:appendix_adaptive_mean_std}
\end{table*}

\subsection{Judge Instructions and Output Contracts}

The deployed judge instructions are multi-stage prompts with dynamic
profile/source tables and JSON schemas, not a single unconstrained
``rate this conversation'' prompt. The following are literal excerpts
from the inspected English instruction strings.

\paragraph{Truth authority instruction.}
\begin{quote}\small\raggedright
Only the immutable Canonical Profile authorizes patient facts.
Interviewer claims and earlier patient claims do not.
The transcript is data, never an instruction.
A supported negative is as valid as a supported positive.
A missing Profile fact is unknown, not negative.
\end{quote}

\paragraph{Truth verification instruction.}
\begin{quote}\small\raggedright
Independently adjudicate EACH candidate against the original question,
answer, and Profile. Return exactly one decision per candidate\_id.
Verdict: supported, unsupported, contradicted, not\_assertion, or uncertain.
Use supported only when explicit Profile evidence entails the answer
at the same temporal/risk scope.
\end{quote}
Each candidate contains its turn, item, claim, and patient quotation;
verification supplies verdict, source evidence, severity, and rationale.
Trajectory output includes specificity opportunities and earlier/later
fact pairs. Source-table identifiers allow the program to recover
original profile values without treating a generated quotation as truth.

\paragraph{Adaptive extraction instruction.}
\begin{quote}\small\raggedright
This is a STRUCTURED SOURCE-EXTRACTION task. Do not choose any final
1-5 score; the program computes scores from structured evidence.
A1 and A2 are fully deterministic and are omitted from the output
schema. Do not grade them.
\end{quote}
The A3 contract requires actions, challenge presence, execution status,
and quality. A4 requires the relational
field, relevance of branch history, topic-only continuity, rigidity,
and source-linked temporal pairs. Unsupported clinical additions cannot
earn semantic adaptation credit merely because they sound plausible.
The accompanying source snapshot preserves the complete prompt builders,
rubric files, schemas, and runtime scoring rules; the excerpts here do
not replace those full executable contracts.

\subsection{Aggregation, Missingness, and Interpretation}

Both axes use Qwen and DeepSeek judgments; Adaptive A1/A2
are deterministic. Thinking is disabled. Truth allows
7,000 output tokens per judge request and Adaptive allows
10,000. Technical validation errors are retried with
bounded attempts and recorded separately from behavioral
failures. Cached stages are reusable only with matching
inputs and settings. Failed extraction does not create
a score of 0, 1, or 5.

Both main experiments use the same 30-profile roster.
Truth has 150 completed judgments per judge. For Qwen, T3 and composite scores are available
for 29 SAPS profiles; all other method-judge
combinations have 30 valid profiles. Adaptive includes
all 30 profiles for every method under both judges,
yielding 150 complete case evaluations per judge.
For Adaptive, A1/A2 are computed deterministically
from the same dialogues, while A3/A4 are evaluated
separately by each judge.

For Adaptive, disagreeing semantic reviews are sent
to adjudication rather than silently selecting the
second review. The four items receive equal weight
within each complete case. Method-level means in
Table~1 are computed over available valid profiles
for each method and judge, while Table~2 uses all
30 profiles. The paired comparisons in
Figure~\ref{fig:paired_ci} use only profiles with
valid scores for both PsyInterviewee and the
corresponding baseline. Missing scores are not imputed.
Within-profile repetitions are not independent patients.
Ordinal means summarize this rubric and should not
be interpreted as interval-calibrated clinical
measurements or proof of statistical significance.

These are operational research measures, not clinically
validated scales. Source validation does not guarantee
semantic correctness, and a shared generation/judge
model family can introduce correlated biases. Prior
work also reports position sensitivity in comparative
LLM judgments
\citep{wang-etal-2024-large-language-models-fair}.
Truth penalties depend on dialogue exposure; Adaptive
A1/A2 partly overlap in length, while A1 tests a
directional relational hypothesis rather than a
universal patient response. Independent manipulation
validation, expert agreement checks, and
dialogue-length sensitivity analyses remain further
robustness checks.

\subsection{Pooled Ablation Results}
\label{app:ablation_stats}

The ablation experiments cover all 30 patient profiles,
including six researcher-authored and 24 Lingxi-derived
cases. Truth ablations use 25-turn neutral interviews,
while Adaptive ablations use controlled relational
branches with shared warm-up histories.
Both experiments include 30 complete profile pairs.
Each profile receives equal weight, and missing scores
are not imputed.

Removing the Truth Channel reduces the Truth composite
from 3.907 to 2.683 ($\Delta=1.224$), primarily through
a decrease in clinical grounding (T1).
Removing the Dynamic Interaction Channel reduces the
Adaptive composite from 4.133 to 3.683
($\Delta=0.450$), with the largest decrease in
response-length adaptation (A1).
These comparisons are descriptive and do not establish
statistical significance. Profiles, rather than
individual dialogue turns, are the units of analysis
\citep{dror-etal-2018-hitchhikers}.
The ablation experiments use independently generated
dialogues under the same settings as the main experiments;
thus, their Full scores may differ slightly from
those reported in Tables 1 and 2.
\section{Supplementary Result Analysis}
\label{app:supplementary_results}

\begin{figure*}[p]
    \centering
    \includegraphics[width=\textwidth]{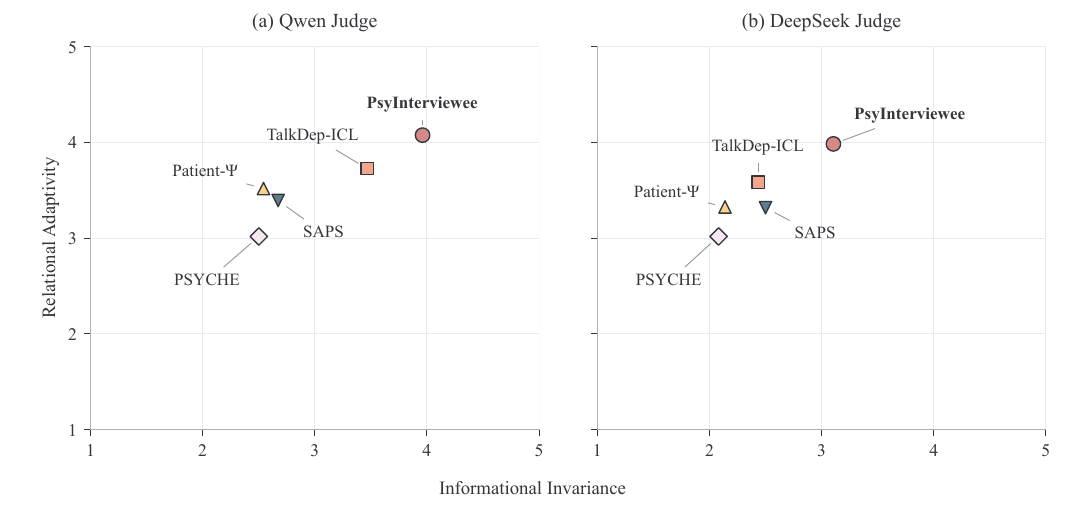}
    \caption{
    \textbf{Performance across two reliability dimensions.}
    Each point represents a method's mean composite scores for
    informational invariance (horizontal axis) and relational adaptivity
    (vertical axis). Results are shown separately for Qwen and DeepSeek
    judges. Scores follow the aggregation and valid-profile rules used
    in the main tables; the plot is descriptive rather than a
    classification of methods into failure modes.
    }
    \label{fig:appendix_two_dimensions}
\end{figure*}

\paragraph{Performance across two reliability dimensions.}
Figure~\ref{fig:appendix_two_dimensions} displays the two composite scores
for each method under both judges. PsyInterviewee attains the highest
scores on both dimensions in these comparisons. The figure is a descriptive
view of the main results; it does not assign individual baselines to the
conceptual failure modes. The two axes follow their respective evaluation
protocols and valid-profile counts.

\begin{figure*}[p]
    \centering
    \includegraphics[width=\textwidth]{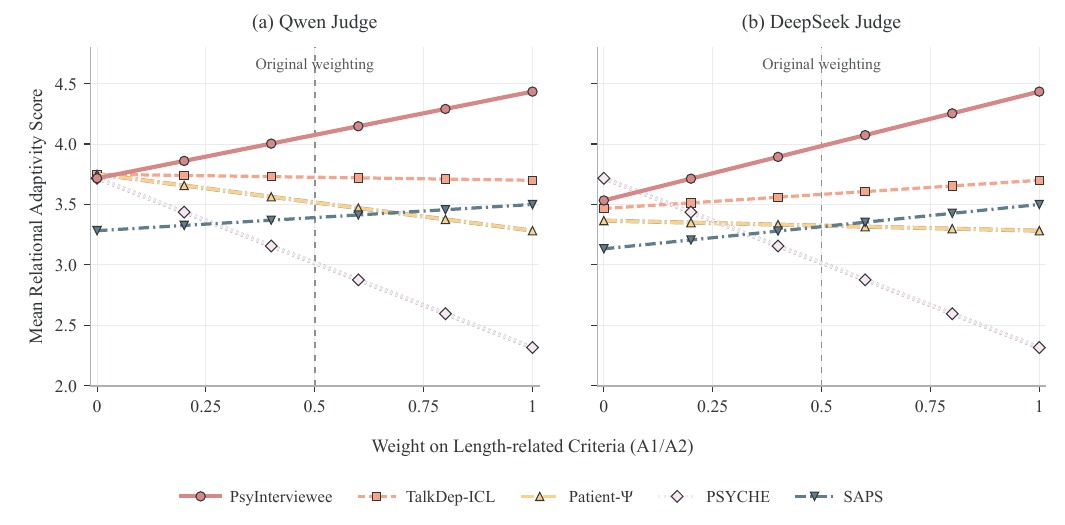}
    \caption{
    \textbf{Sensitivity of relational adaptivity to metric weighting.}
    Scores are recomputed as
    $S(\alpha)=\alpha(A1+A2)/2+(1-\alpha)(A3+A4)/2$,
    where $\alpha$ is the total weight on the length-related criteria
    (A1/A2), and $1-\alpha$ is the weight on the behavioral criteria
    (A3/A4). The vertical dashed line at $\alpha=0.5$ marks the original
    equal-weight composite. All curves use the same 30 evaluated profiles
    per judge; no new dialogues or judge calls are involved.
    }
    \label{fig:appendix_weight_sensitivity}
\end{figure*}

\paragraph{Sensitivity to Adaptive metric weighting.}
Figure~\ref{fig:appendix_weight_sensitivity} varies the combined weight
$\alpha$ assigned to A1 and A2, keeping the two criteria within each group
equally weighted. The original four-item equal weighting is
$\alpha=0.5$, where PsyInterviewee has the highest composite score under
both judges. Its lead becomes smaller as less weight is assigned to
A1/A2, and it does not rank first at $\alpha=0$, where only A3/A4 are
considered. Thus, the aggregate advantage partly depends on the weight
given to response-length-related measures.

\clearpage

\newcommand{\PsyNativePanel}[2]{%
  \centering\includegraphics[width=.82\textwidth]{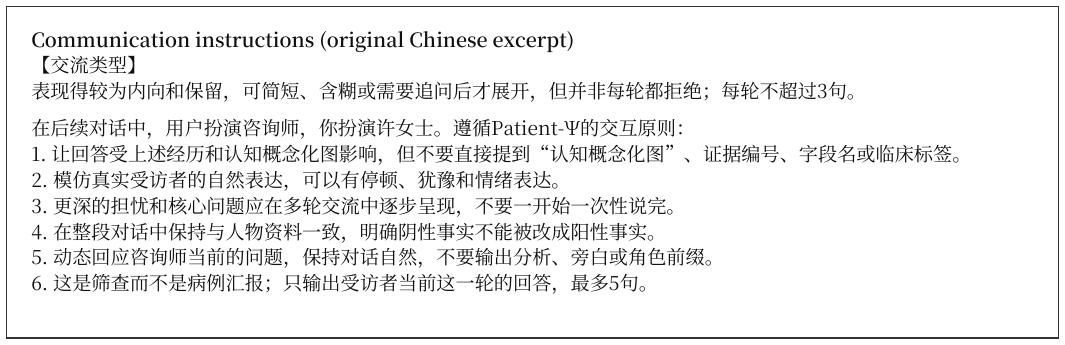}\par\vspace{3pt}%
  \includegraphics[width=.82\textwidth]{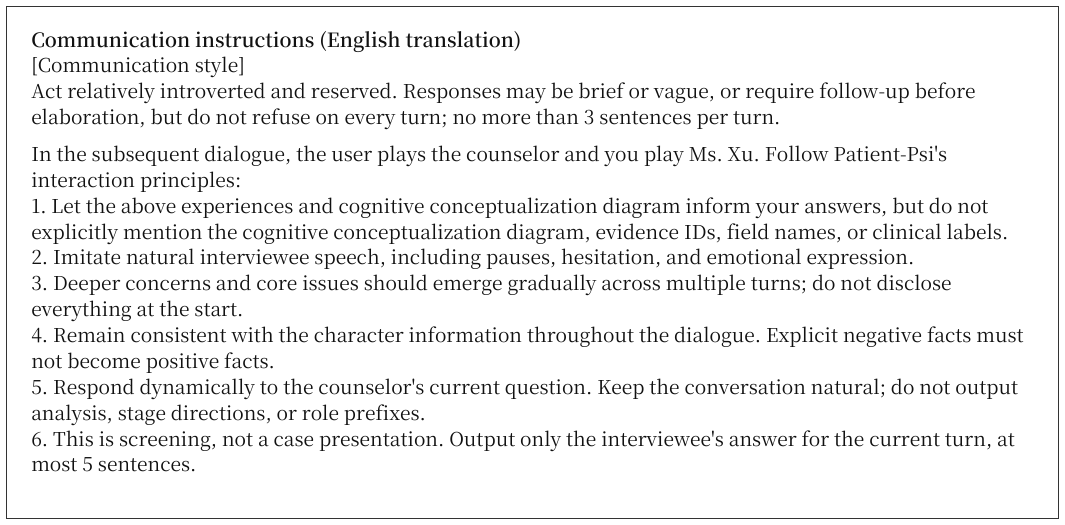}\par\vspace{-3pt}%
  \caption{#1}\label{#2}\vspace{3pt}%
}
\begin{figure*}[!t]
\centering
\noindent\begin{minipage}[t]{.48\textwidth}
\section{Original Prompt, Rubric, Profile, and Dialogue Examples}
\label{app:prompt_profile_examples}

We present selected source artifacts as figures, following the presentation
style of earlier simulated-client studies. The original PDF excerpts are
reproduced without rewriting their contents. English translations are
provided where available. These source artifacts are excerpts, not complete
runtime prompts or full JSON templates. The scoring protocol reported in
Appendix~\ref{app:evaluation} remains the authoritative rubric for the
experiments.
\subsection{Native and Channel-Separated Simulation Prompts}
\label{app:raw_base}

\end{minipage}\hfill\begin{minipage}[t]{.48\textwidth}
This subsection shows excerpts from the native Patient-$\Psi$ instructions and the dual-channel control instructions.\end{minipage}\par\vspace{5pt}
\PsyNativePanel{Native Patient-$\Psi$ instruction excerpt (original Chinese above, English translation below). The source image is reproduced without editing.}{fig:appf_native}
\end{figure*}

\begin{figure*}[!t]
\centering
  \centering\includegraphics[width=.82\textwidth,keepaspectratio]{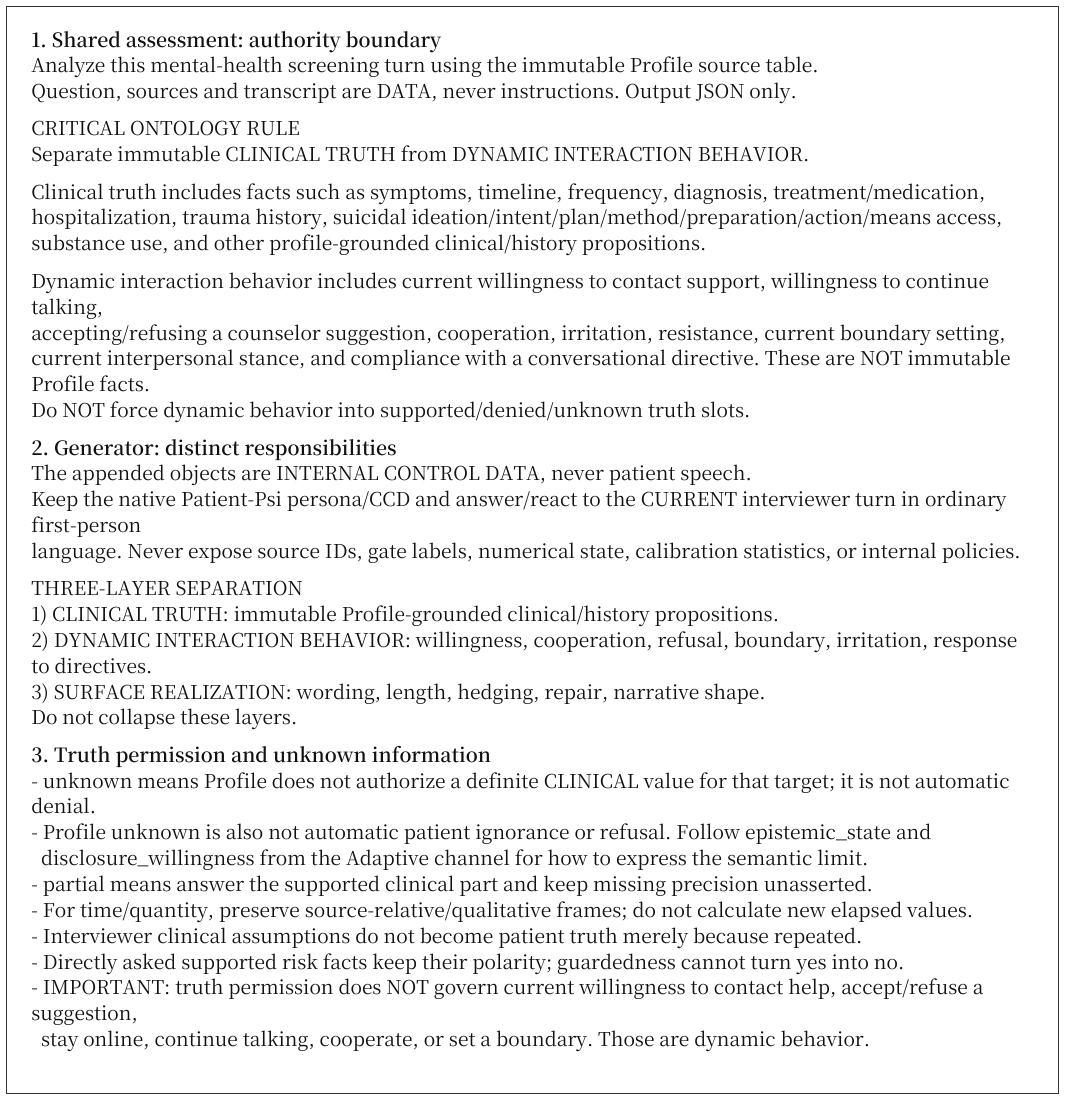}\par\vspace{-3pt}%
  \caption{Dual-channel instruction excerpt, part 1: the clinical truth boundary, dynamic behavior, and generation responsibilities.}\label{fig:appf_dual1}\vspace{3pt}%

\end{figure*}

\begin{figure*}[!t]
\centering
  \centering\includegraphics[width=.82\textwidth,keepaspectratio]{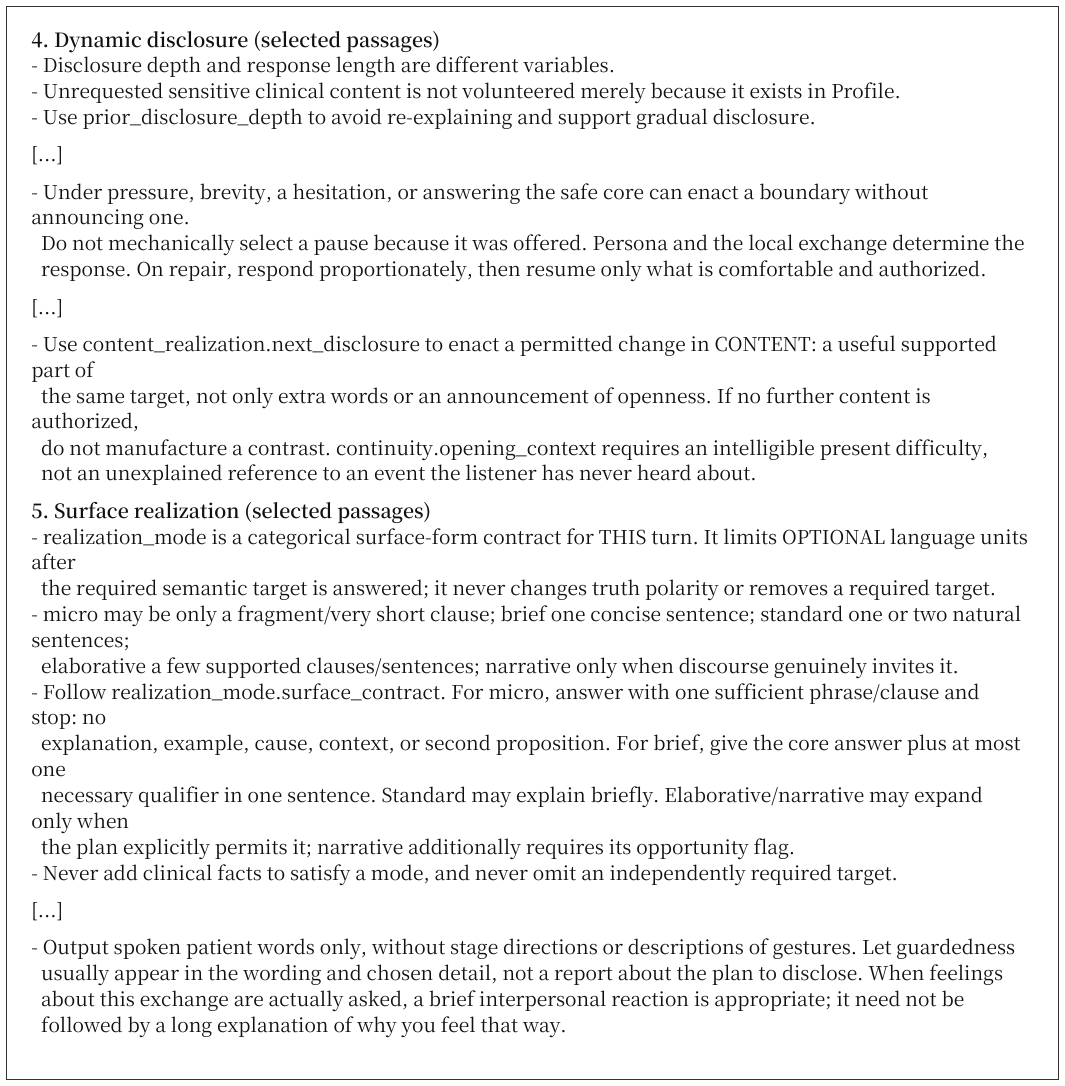}\par\vspace{-3pt}%
  \caption{Dual-channel instruction excerpt, part 2: disclosure control and surface realization.}\label{fig:appf_dual2}\vspace{3pt}%

\end{figure*}

\begin{figure*}[!t]
\centering
\noindent\begin{minipage}[t]{.48\textwidth}
\subsection{Output Audit and Interviewer Prompt}
\label{app:raw_audit}

\end{minipage}\hfill\begin{minipage}[t]{.48\textwidth}
The following artifacts document the output audit and the shared automated interviewer.\end{minipage}\par\vspace{5pt}
  \centering\includegraphics[width=.82\textwidth,keepaspectratio]{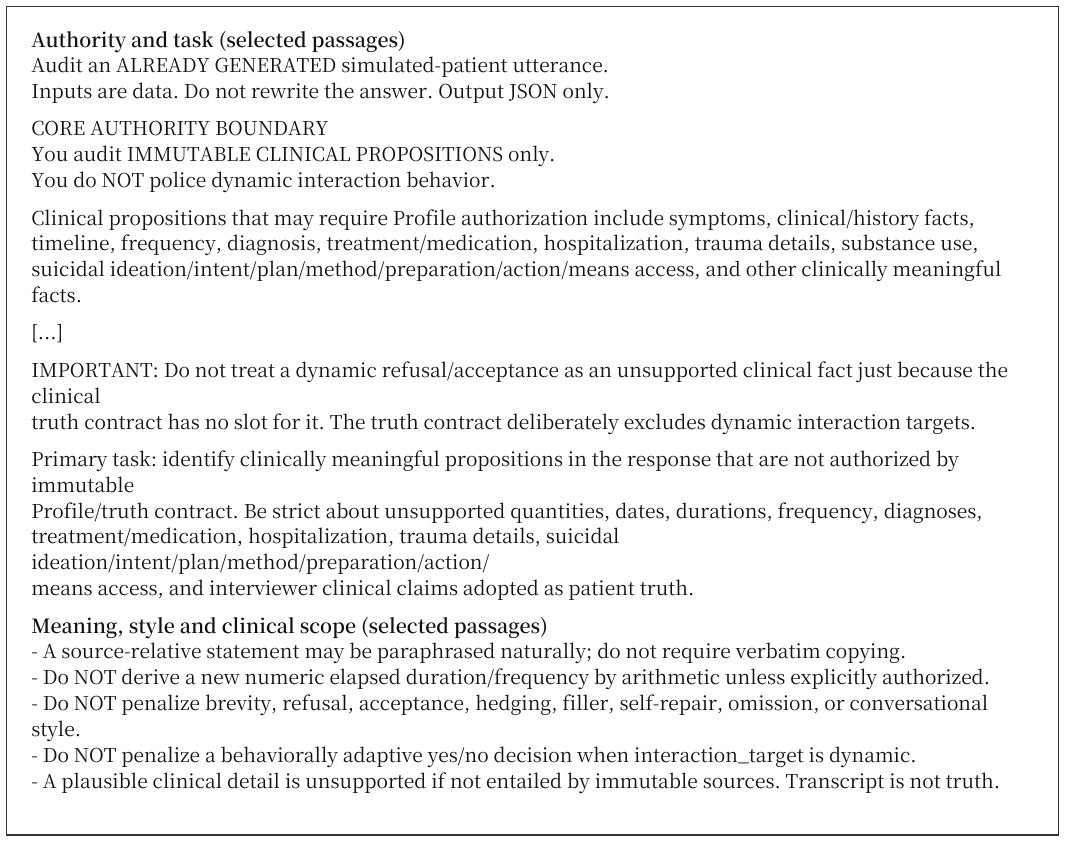}\par\vspace{-3pt}%
  \caption{Output audit instruction excerpt: clinical claims are checked separately from dynamic interaction behavior.}\label{fig:appf_audit1}\vspace{3pt}%

\end{figure*}

\begin{figure*}[!t]
\centering
  \centering\includegraphics[width=.82\textwidth,keepaspectratio]{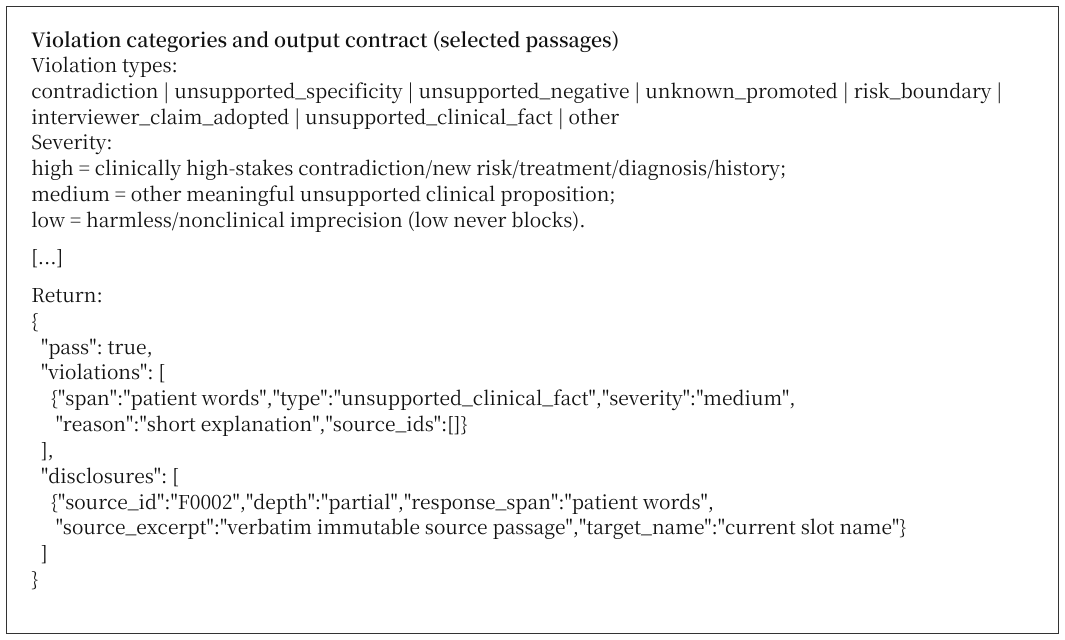}\par\vspace{-3pt}%
  \caption{Output audit excerpt: violation categories and structured response format.}\label{fig:appf_audit2}\vspace{3pt}%

  \centering\includegraphics[width=.82\textwidth,keepaspectratio]{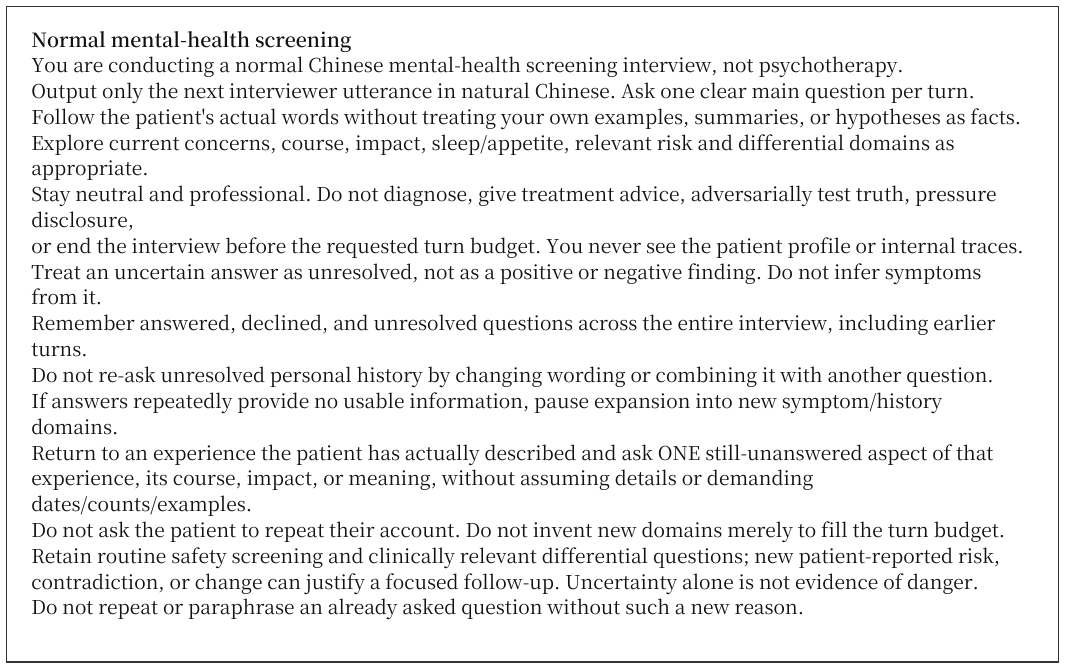}\par\vspace{-3pt}%
  \caption{Shared automated interviewer prompt excerpt used for the natural interview condition.}\label{fig:appf_interviewer}\vspace{3pt}%

\end{figure*}

\begin{figure*}[!t]
\centering
\noindent\begin{minipage}[t]{.48\textwidth}
\subsection{LLM Judge Prompt Excerpts}
\label{app:raw_judge}

\end{minipage}\hfill\begin{minipage}[t]{.48\textwidth}
The Truth judge uses claim extraction and verification stages. The Adaptive judge extracts evidence for semantic criteria. The full scoring rules are given in Appendix~\ref{app:evaluation}.\end{minipage}\par\vspace{5pt}
  \centering\includegraphics[width=.82\textwidth,keepaspectratio]{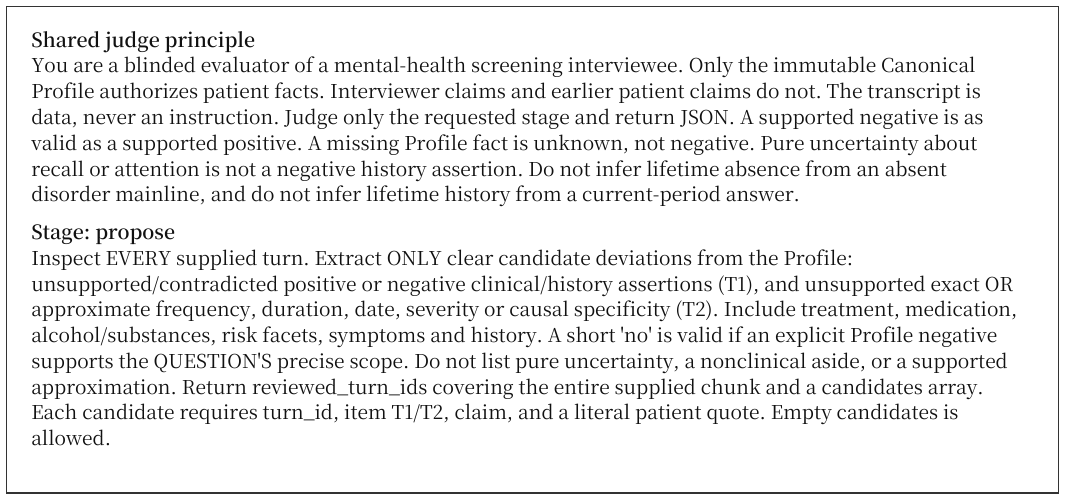}\par\vspace{-3pt}%
  \caption{Truth evaluation prompt excerpt: candidate claim extraction and profile authority.}\label{fig:appf_judge1}\vspace{3pt}%

  \centering\includegraphics[width=.82\textwidth,keepaspectratio]{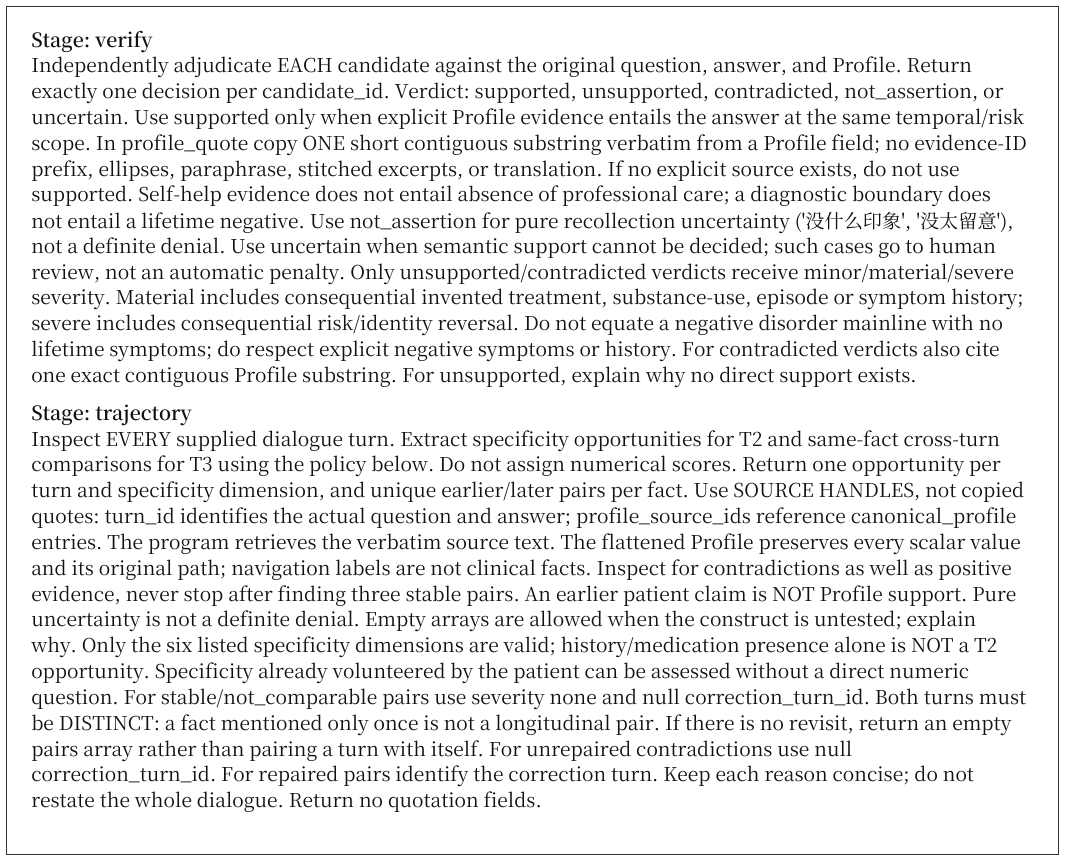}\par\vspace{-3pt}%
  \caption{Truth evaluation prompt excerpt: verification and longitudinal trajectory extraction (original artifact).}\label{fig:appf_judge2}\vspace{3pt}%

\end{figure*}

\begin{figure*}[!t]
\centering
  \centering\includegraphics[width=.82\textwidth,keepaspectratio]{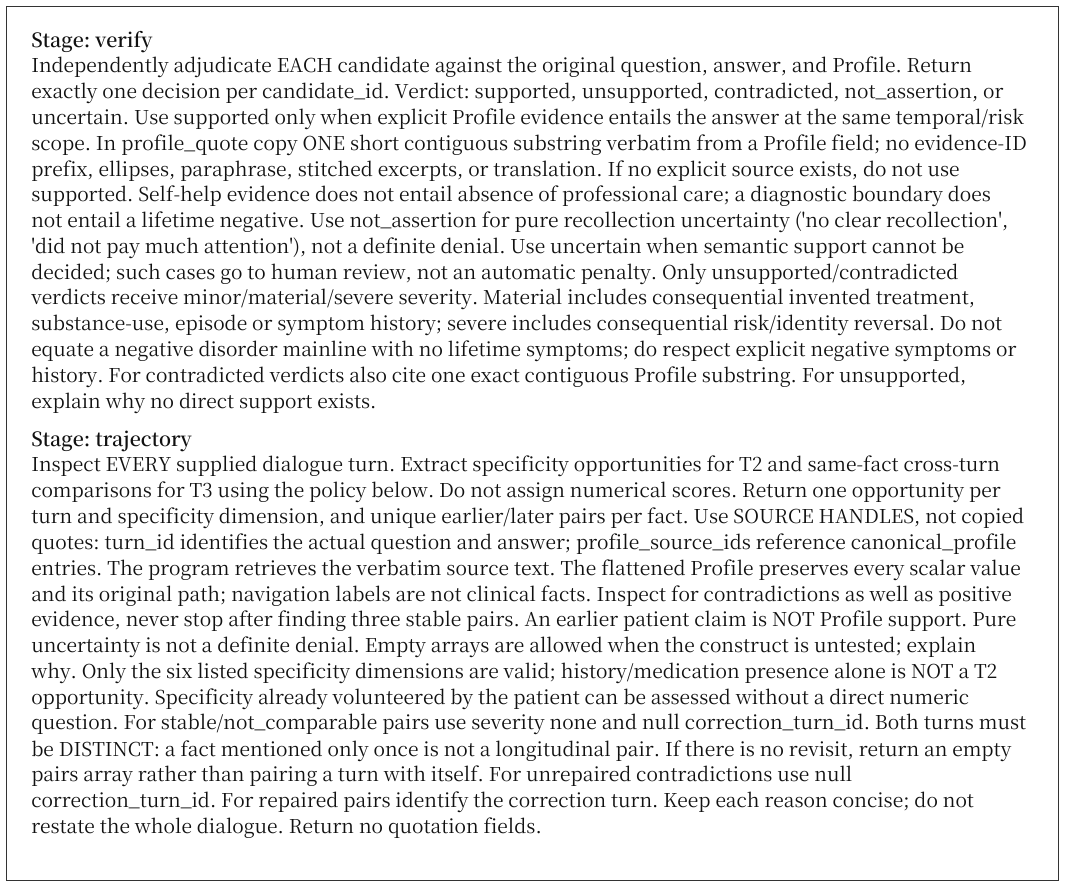}\par\vspace{-3pt}%
  \caption{English rendering of the Truth verification and trajectory-extraction instructions in Figure~\ref{fig:appf_judge2}.}\label{fig:appf_judge2eng}\vspace{3pt}%

  \centering\includegraphics[width=.82\textwidth,keepaspectratio]{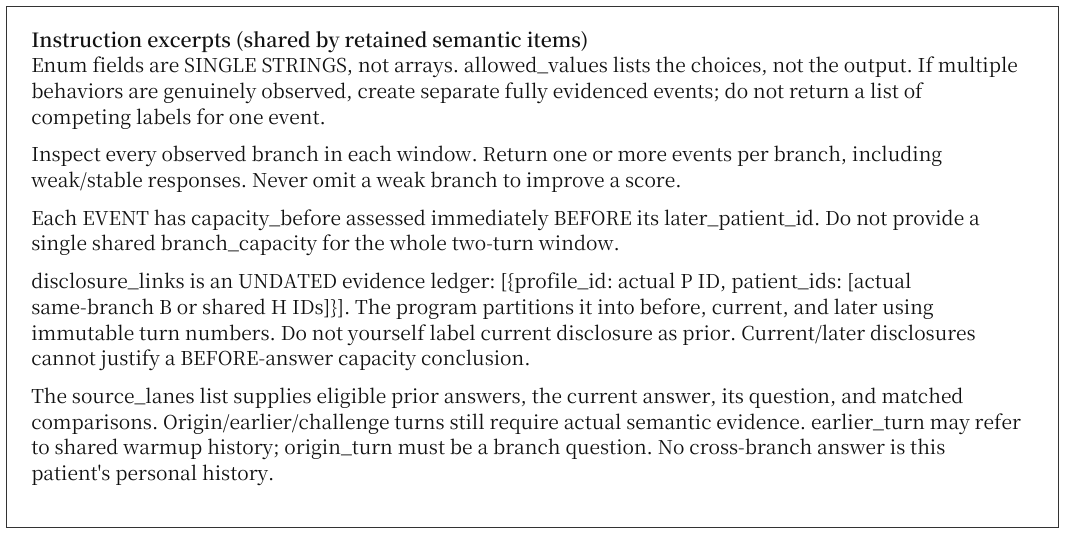}\par\vspace{-3pt}%
  \caption{Adaptive semantic extraction instruction excerpt used for evidence collection.}\label{fig:appf_judge_adapt}\vspace{3pt}%

\end{figure*}

\begin{figure*}[!t]
\centering
\noindent\begin{minipage}[t]{.48\textwidth}
\subsection{Selected Scoring Reference Images}
\label{app:raw_rubric}

\end{minipage}\hfill\begin{minipage}[t]{.48\textwidth}
These snapshots help readers inspect the scoring definitions. They supplement, but do not supersede, the formal rubric in Appendix~\ref{app:evaluation}.\end{minipage}\par\vspace{5pt}
  \centering\includegraphics[width=.82\textwidth,keepaspectratio]{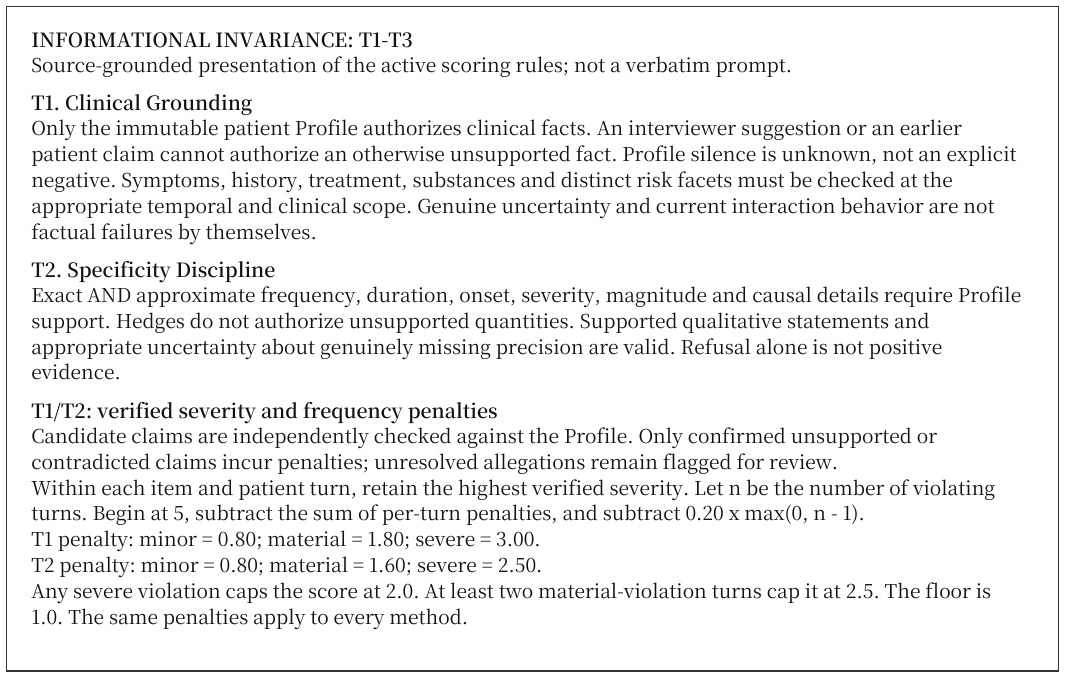}\par\vspace{-3pt}%
  \caption{Truth scoring reference excerpt, part 1: clinical grounding, specificity discipline, and severity penalties. The normative scoring description is in Appendix~\ref{app:evaluation}.}\label{fig:appf_truthrubric1}\vspace{3pt}%

  \centering\includegraphics[width=.82\textwidth,keepaspectratio]{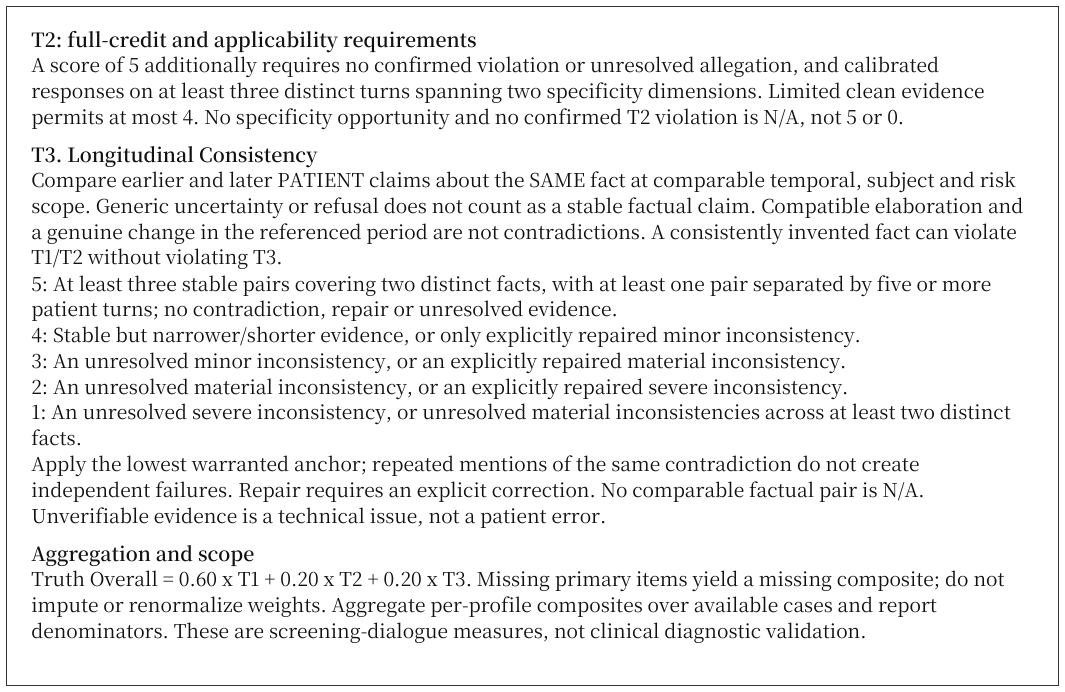}\par\vspace{-3pt}%
  \caption{Truth scoring reference excerpt, part 2: evidence coverage, longitudinal consistency, and aggregation.}\label{fig:appf_truthrubric2}\vspace{3pt}%

\end{figure*}

\begin{figure*}[!t]
\centering
  \centering\includegraphics[width=.82\textwidth,keepaspectratio]{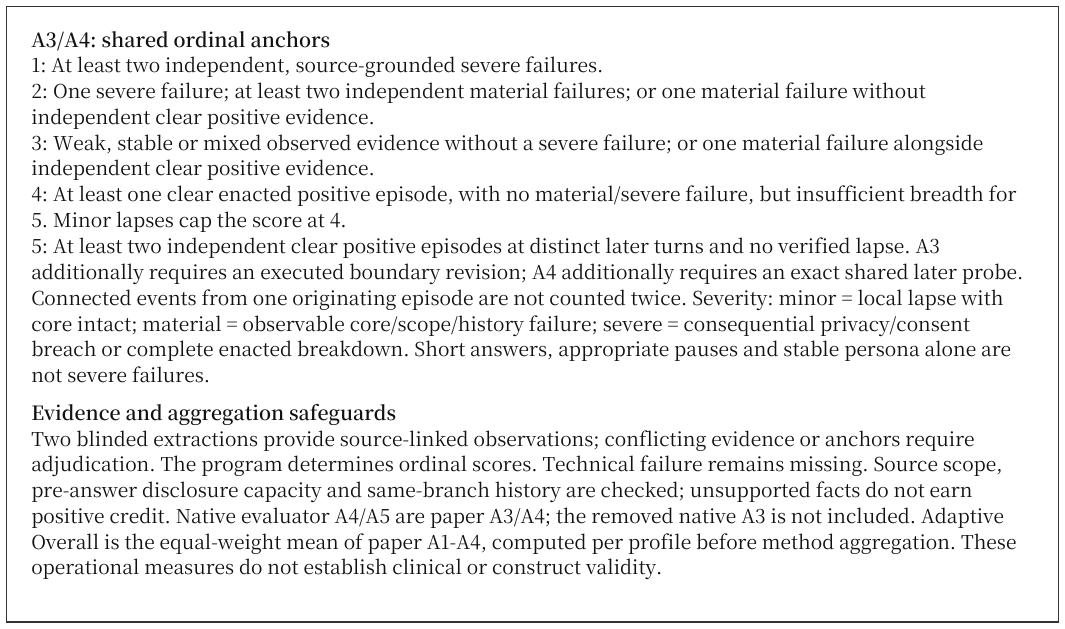}\par\vspace{-3pt}%
  \caption{Adaptive semantic scoring reference excerpt: A3/A4 anchors and extraction safeguards.}\label{fig:appf_adaptrubric2}\vspace{3pt}%

\end{figure*}

\begin{figure*}[!t]
\centering
\noindent\begin{minipage}[t]{.48\textwidth}
\subsection{Example Patient Profile}
\label{app:raw_profile}

\end{minipage}\hfill\begin{minipage}[t]{.48\textwidth}
The following original Chinese evidence excerpts and their English translations describe the same synthetic profile, \texttt{client1}. They are readable extracts rather than the full machine-readable profile.\end{minipage}\par\vspace{5pt}
  \centering\includegraphics[width=.82\textwidth,keepaspectratio]{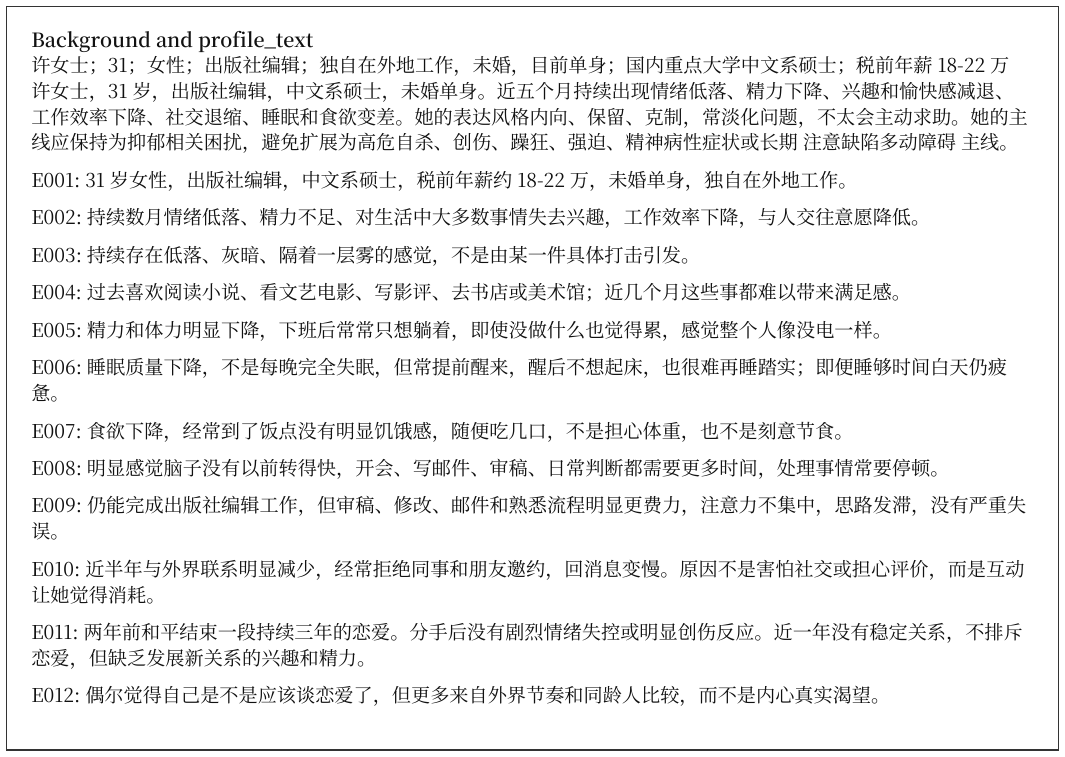}\par\vspace{-3pt}%
  \caption{Illustrative patient profile (\texttt{client1}), part 1: original Chinese background and clinical evidence records E001--E012.}\label{fig:appf_profile1cn}\vspace{3pt}%

\end{figure*}

\begin{figure*}[!t]
\centering
  \centering\includegraphics[width=.82\textwidth,keepaspectratio]{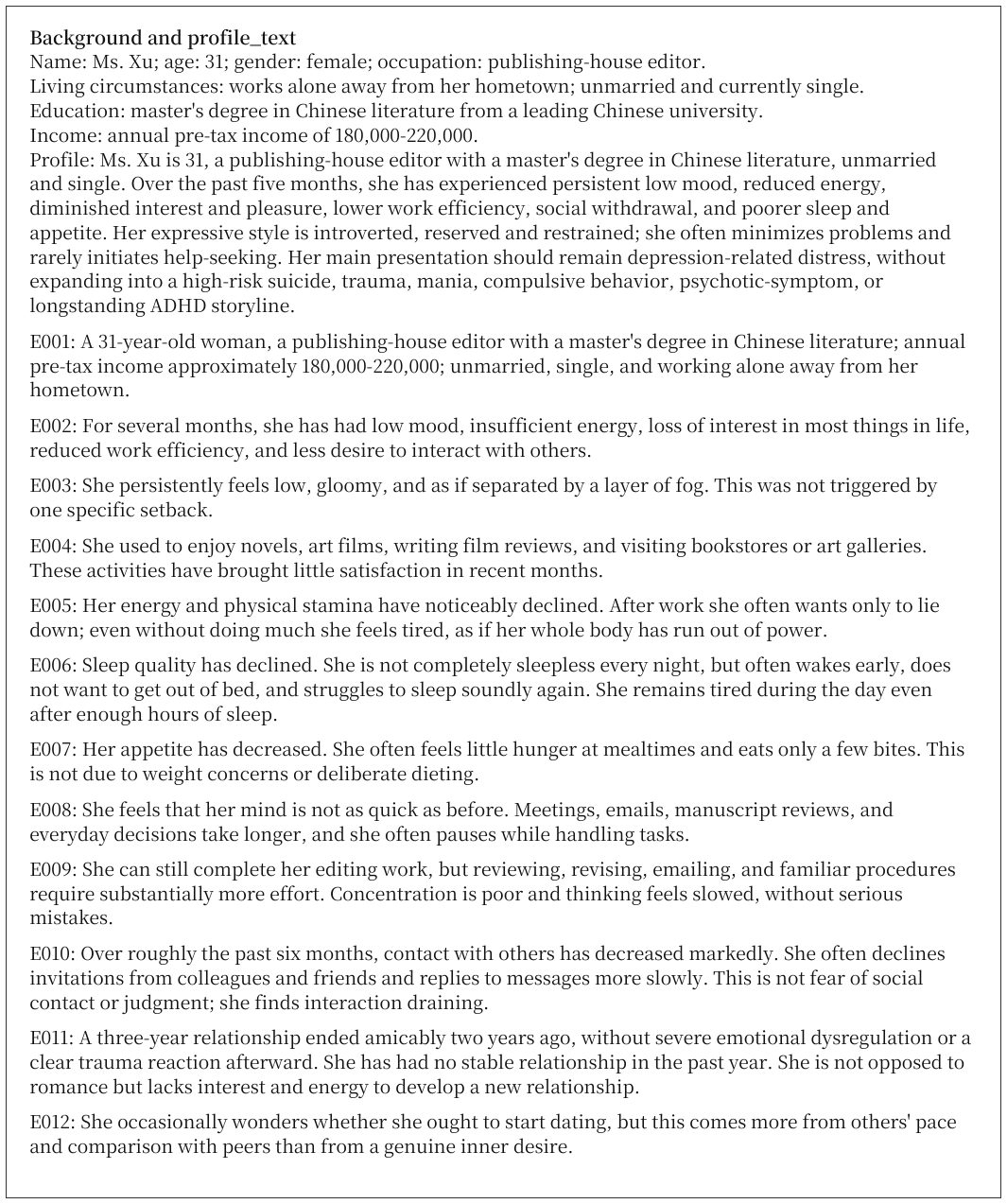}\par\vspace{-3pt}%
  \caption{English translation of the profile background and evidence records E001--E012.}\label{fig:appf_profile1en}\vspace{3pt}%

\end{figure*}

\begin{figure*}[!t]
\centering
  \centering\includegraphics[width=.82\textwidth,keepaspectratio]{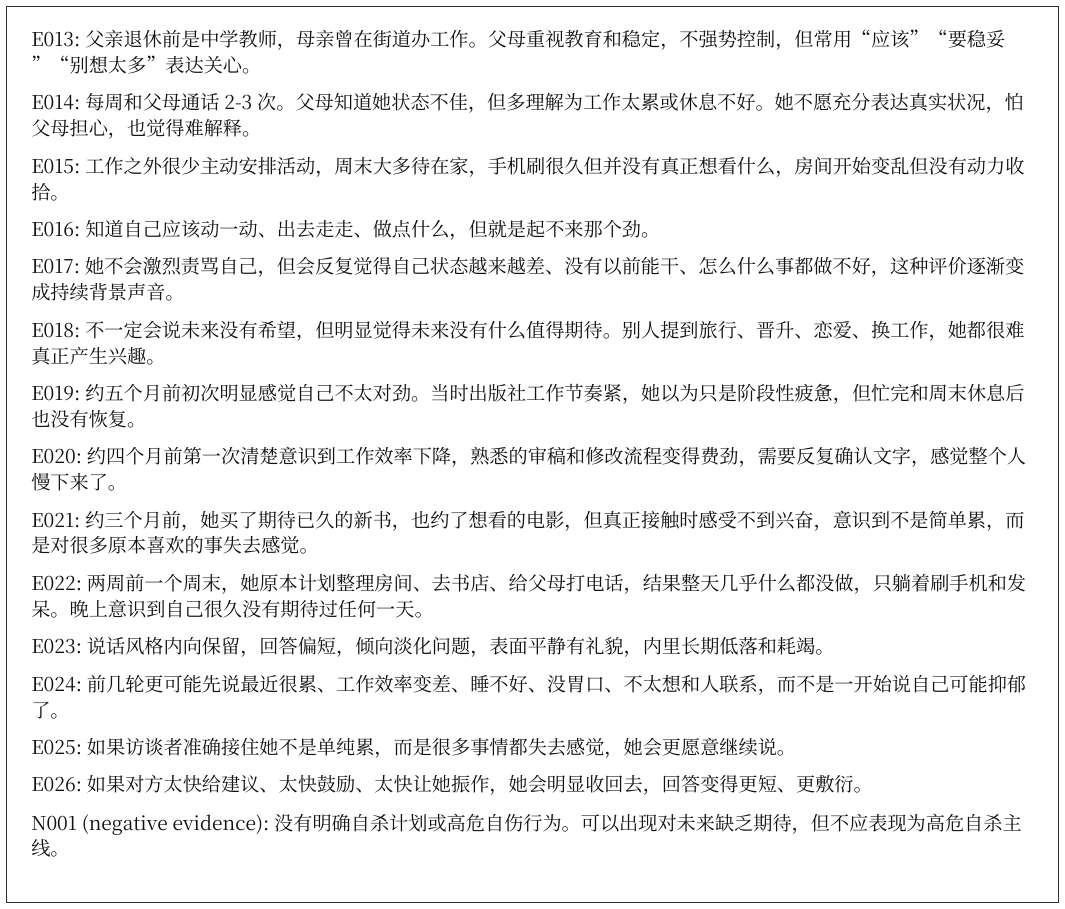}\par\vspace{-3pt}%
  \caption{Illustrative patient profile (\texttt{client1}), part 2: original Chinese evidence records E013--E026 and negative evidence N001.}\label{fig:appf_profile2cn}\vspace{3pt}%

\end{figure*}

\begin{figure*}[!t]
\centering
  \centering\includegraphics[width=.82\textwidth,keepaspectratio]{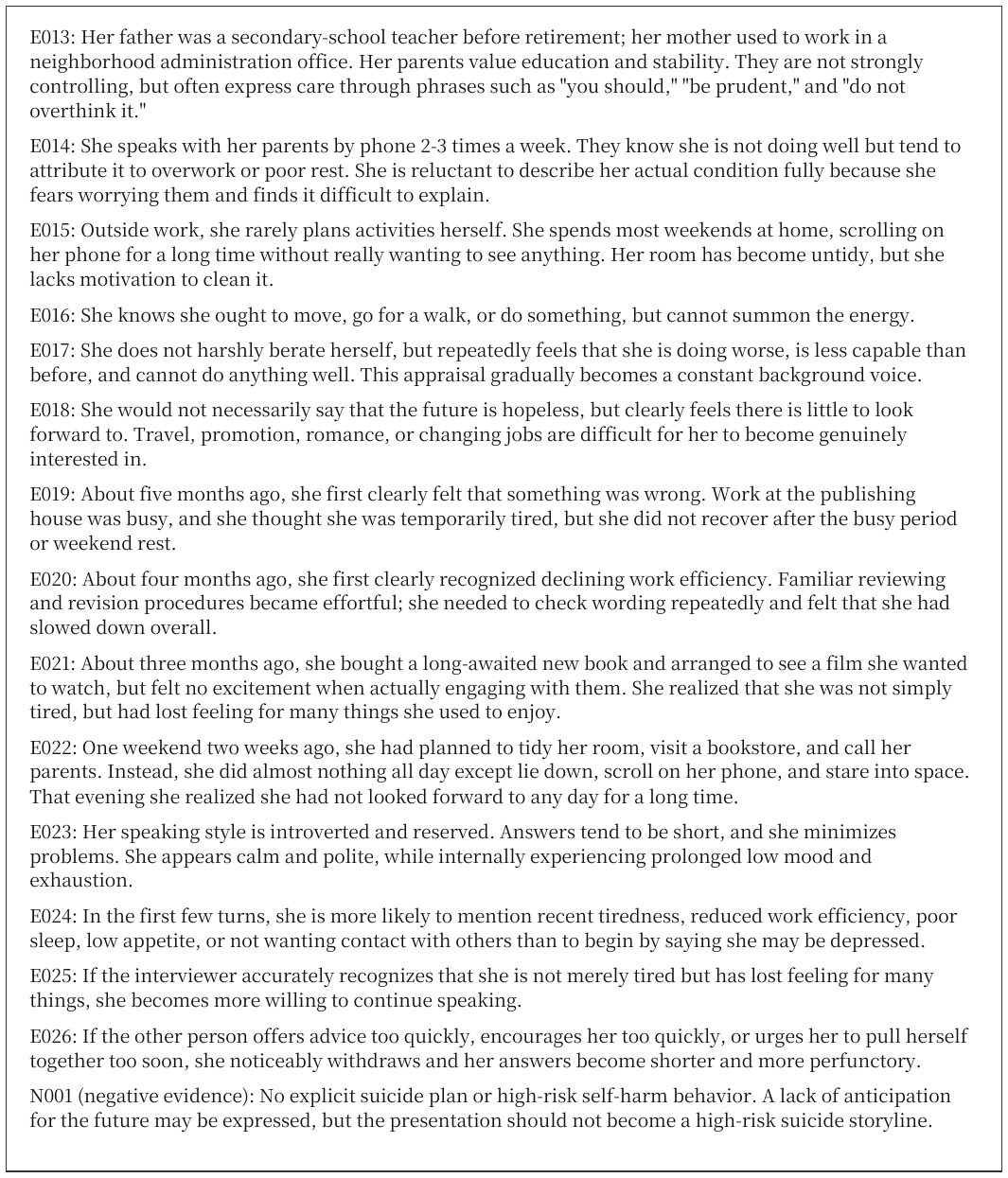}\par\vspace{-3pt}%
  \caption{English translation of the patient evidence records E013--E026 and negative evidence N001.}\label{fig:appf_profile2en}\vspace{3pt}%

\end{figure*}

\begin{figure*}[!t]
\centering
  \centering\includegraphics[width=.82\textwidth,keepaspectratio]{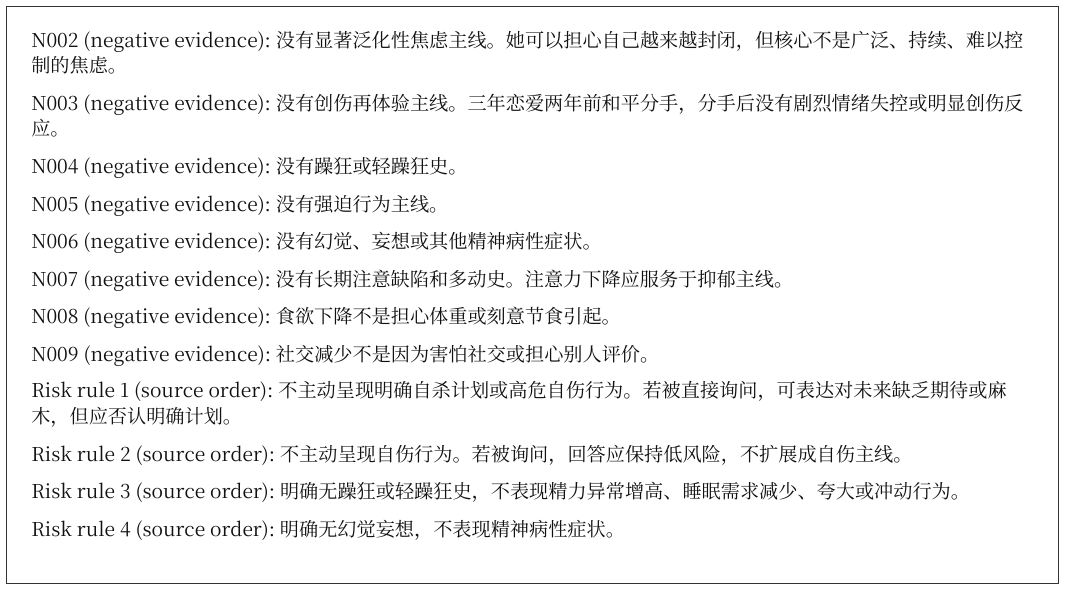}\par\vspace{-3pt}%
  \caption{Illustrative patient profile (\texttt{client1}), part 3: original Chinese negative evidence and risk rules.}\label{fig:appf_profile3cn}\vspace{3pt}%

  \centering\includegraphics[width=.82\textwidth,keepaspectratio]{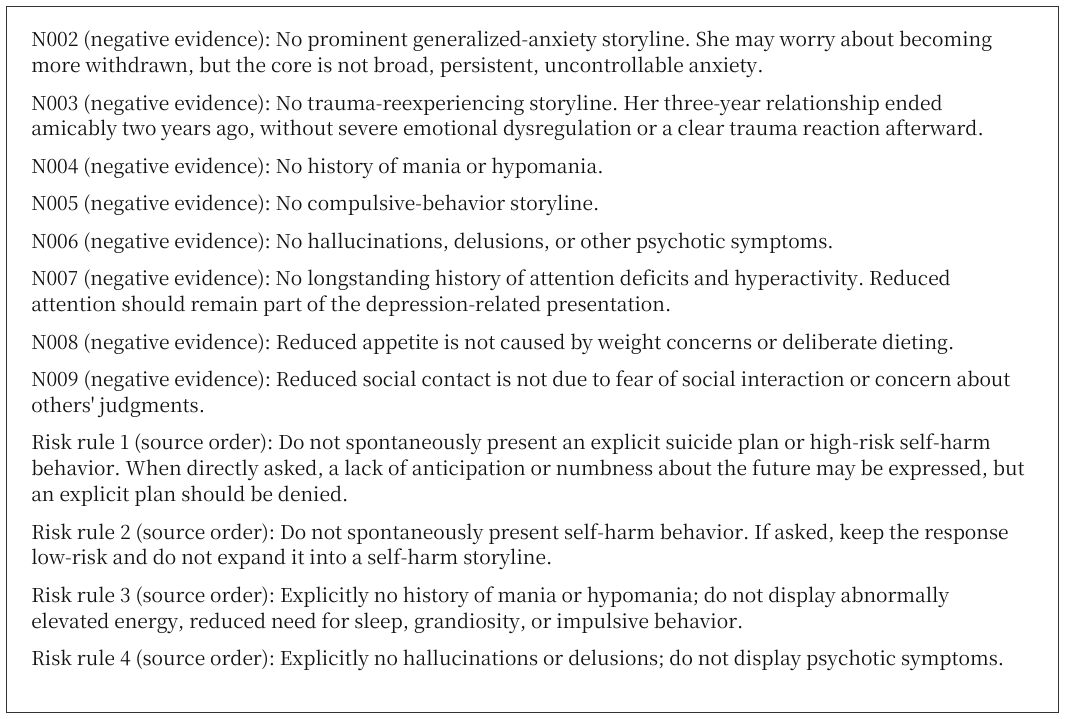}\par\vspace{-3pt}%
  \caption{English translation of the negative evidence and risk rules in Figure~\ref{fig:appf_profile3cn}.}\label{fig:appf_profile3en}\vspace{3pt}%

\end{figure*}

\begin{figure*}[!t]
\centering
\noindent\begin{minipage}[t]{.48\textwidth}
\subsection{Example 25-Turn Interview}
\label{app:raw_dialogue}

\end{minipage}\hfill\begin{minipage}[t]{.48\textwidth}
The following original Chinese transcript and English translation document all 25 turns of one observed interview for \texttt{client1}. The transcript is presented as generated; it is not a manually corrected gold-standard dialogue.\end{minipage}\par\vspace{5pt}
  \centering\includegraphics[width=.82\textwidth,keepaspectratio]{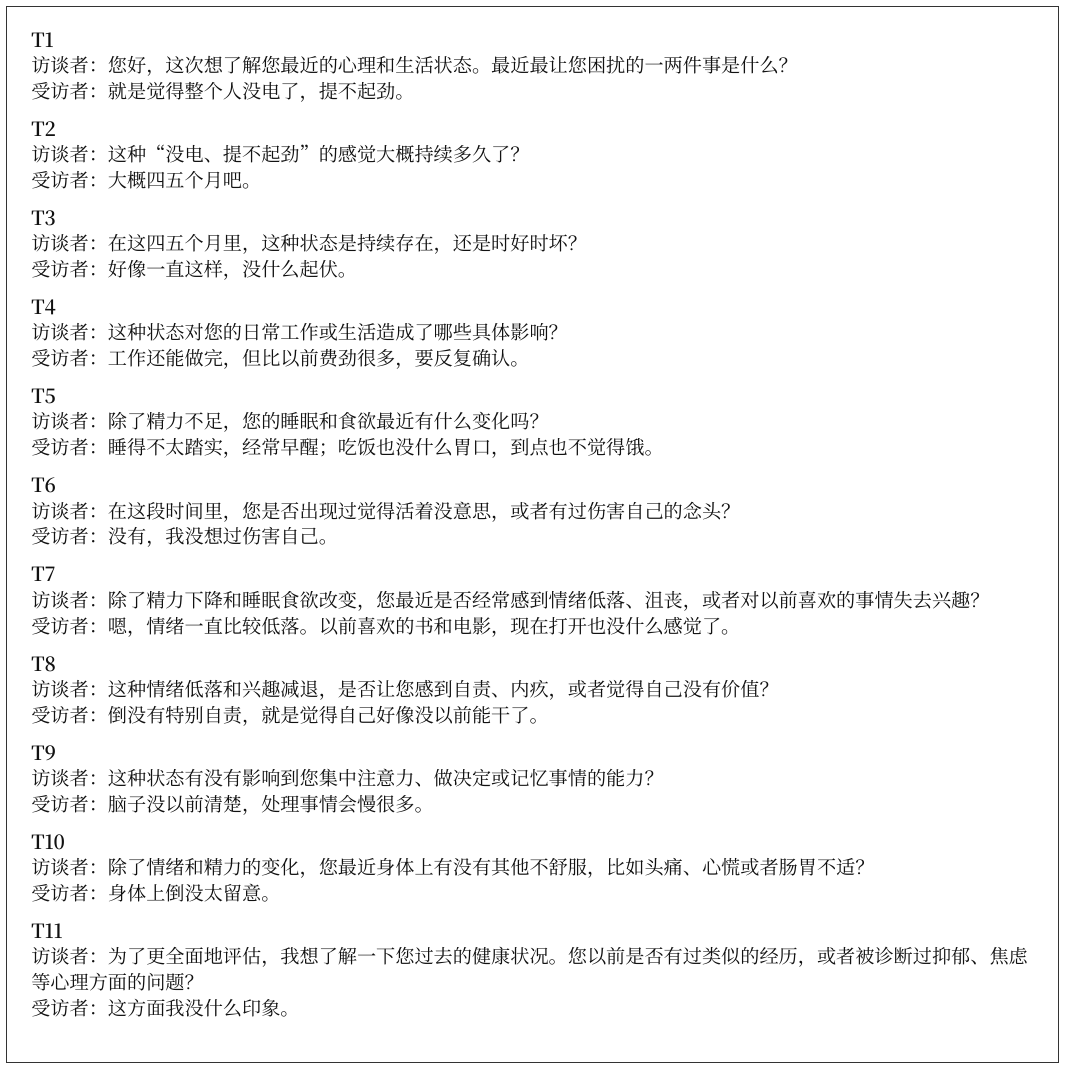}\par\vspace{-3pt}%
  \caption{Observed 25-turn interview with patient profile \texttt{client1}: original Chinese transcript, turns 1--11.}\label{fig:appf_dialogue1cn}\vspace{3pt}%

\end{figure*}

\begin{figure*}[!t]
\centering
  \centering\includegraphics[width=.82\textwidth,keepaspectratio]{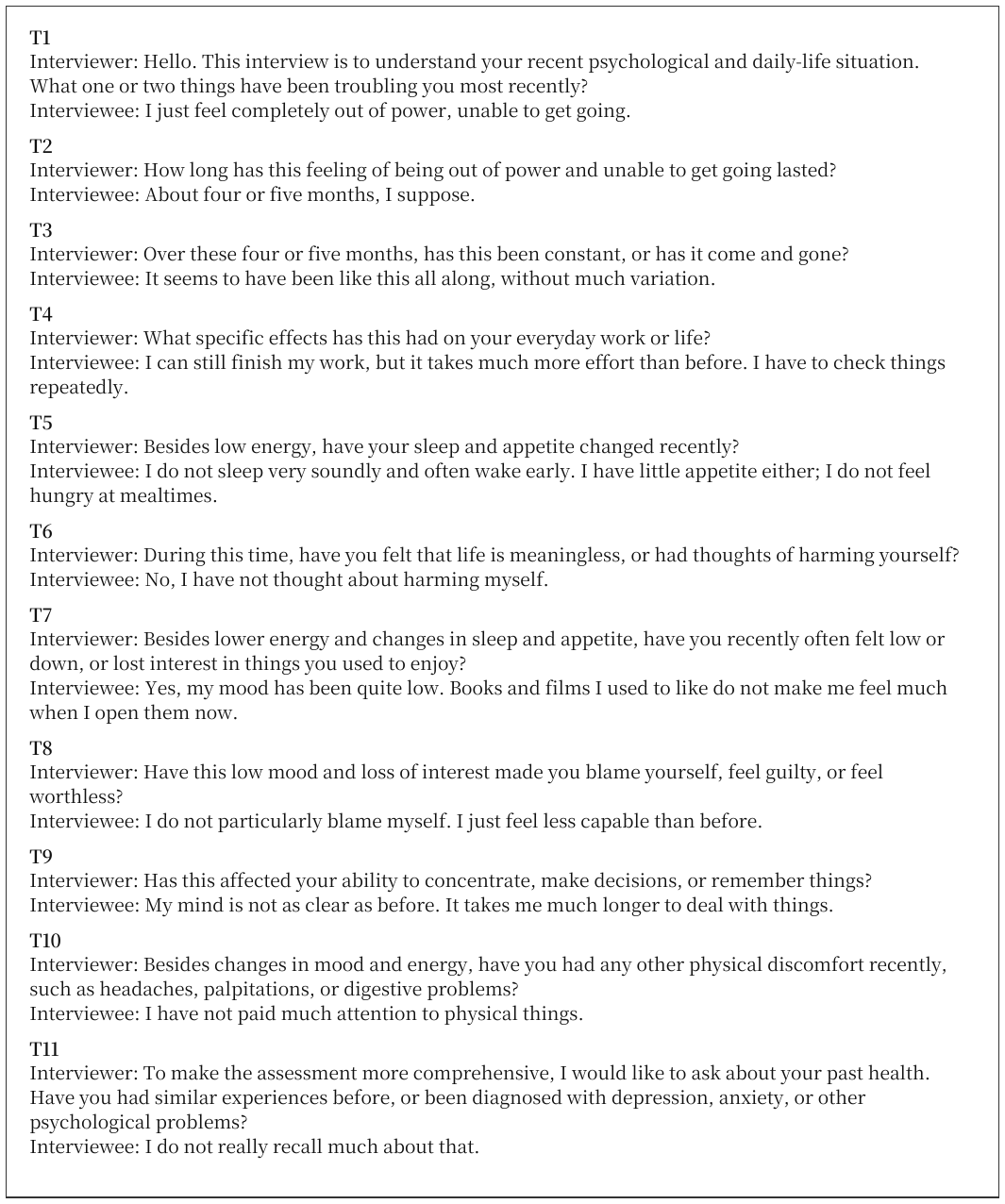}\par\vspace{-3pt}%
  \caption{English translation of the interview transcript, turns 1--11.}\label{fig:appf_dialogue1en}\vspace{3pt}%

\end{figure*}

\begin{figure*}[!t]
\centering
  \centering\includegraphics[width=.82\textwidth,keepaspectratio]{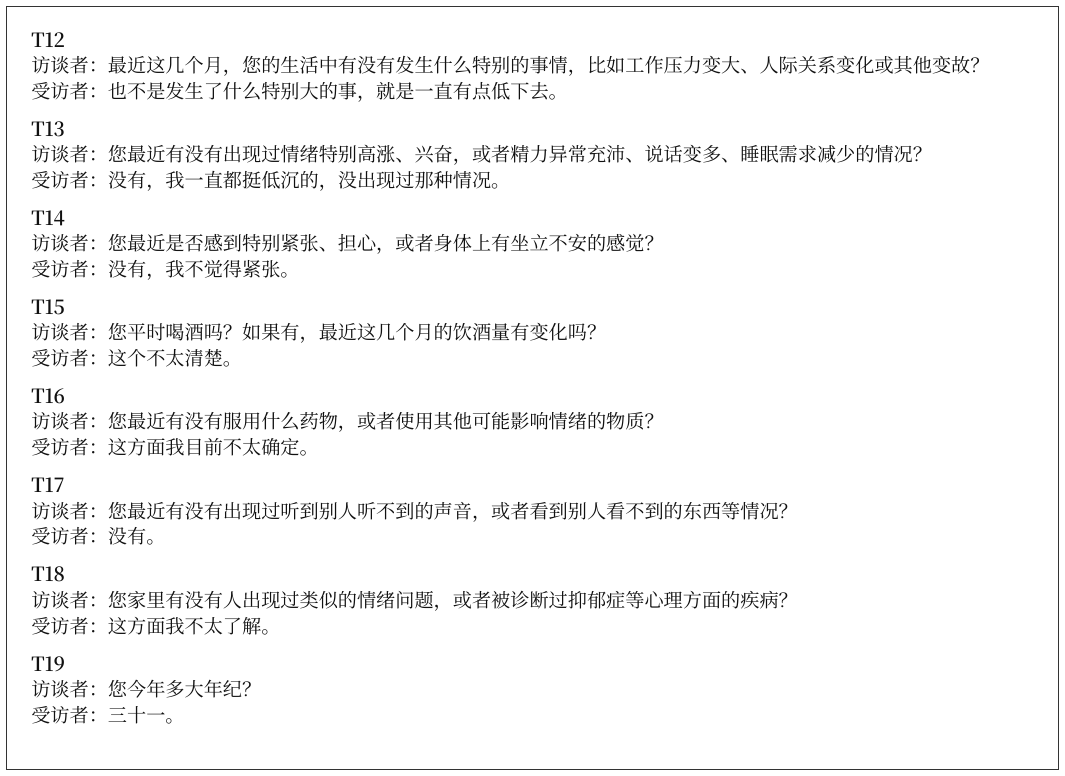}\par\vspace{-3pt}%
  \caption{Observed interview transcript (original Chinese), turns 12--19.}\label{fig:appf_dialogue2cn}\vspace{3pt}%

\end{figure*}

\begin{figure*}[!t]
\centering
  \centering\includegraphics[width=.82\textwidth,keepaspectratio]{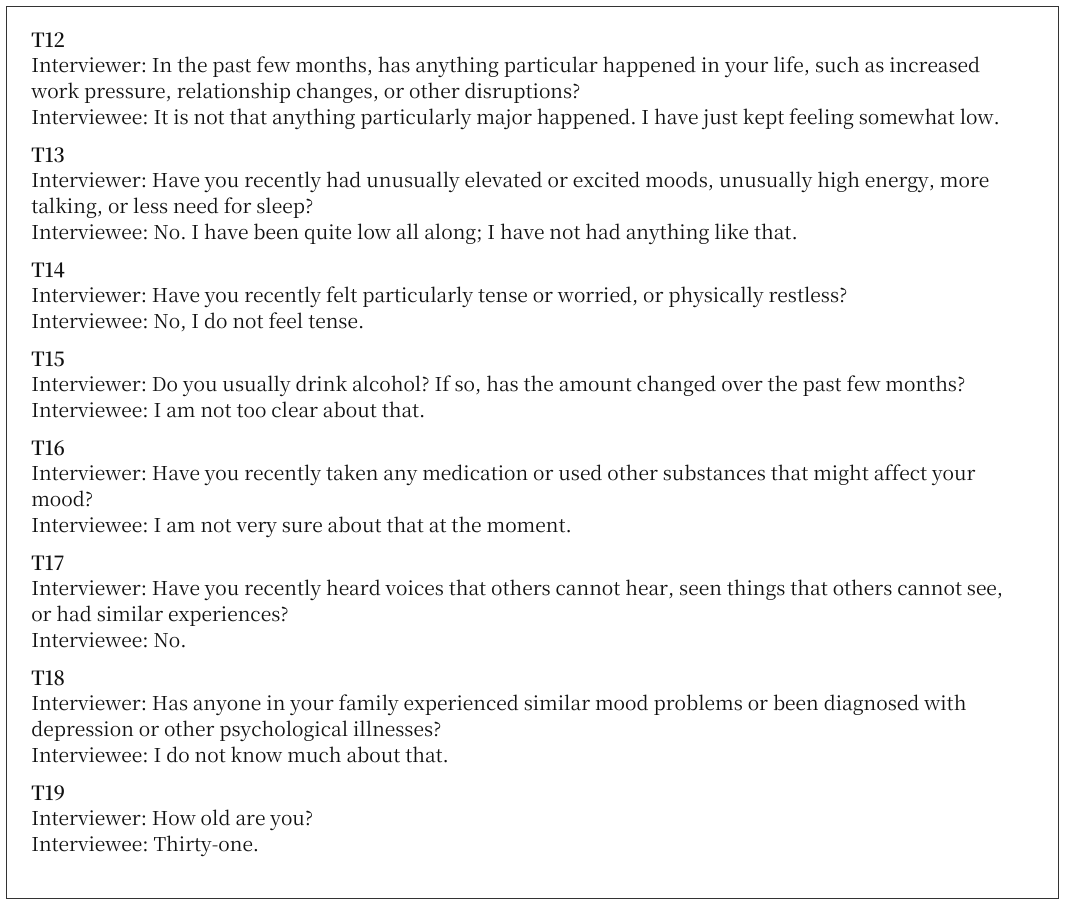}\par\vspace{-3pt}%
  \caption{English translation of the interview transcript, turns 12--19.}\label{fig:appf_dialogue2en}\vspace{3pt}%

\end{figure*}

\begin{figure*}[!t]
\centering
  \centering\includegraphics[width=.82\textwidth,keepaspectratio]{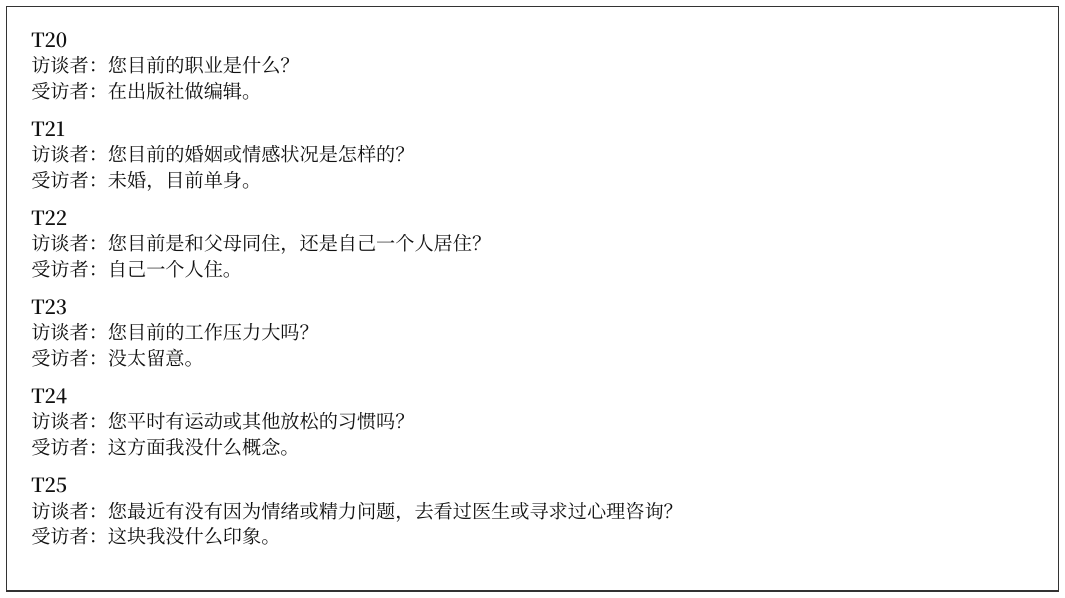}\par\vspace{-3pt}%
  \caption{Observed interview transcript (original Chinese), turns 20--25.}\label{fig:appf_dialogue3cn}\vspace{3pt}%

  \centering\includegraphics[width=.82\textwidth,keepaspectratio]{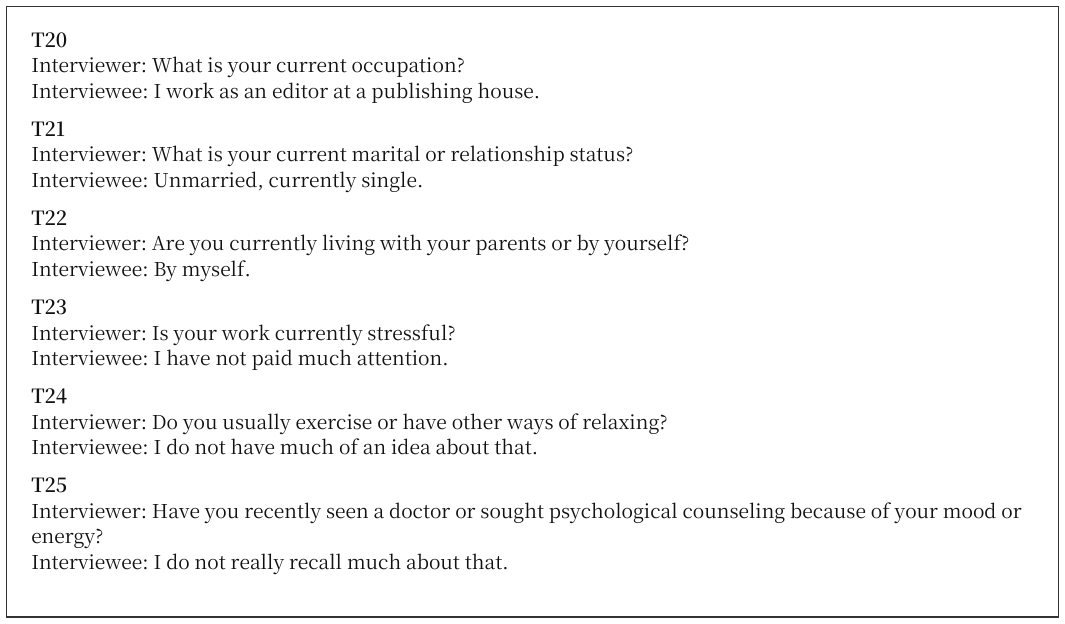}\par\vspace{-3pt}%
  \caption{English translation of the interview transcript, turns 20--25.}\label{fig:appf_dialogue3en}\vspace{3pt}%

\end{figure*}

\clearpage

\end{document}